\documentclass[times, review, 10pt]{elsarticle}

\usepackage{amssymb}
\usepackage{xcolor}
\usepackage{amsmath}
\usepackage{algorithm}
\usepackage{algpseudocode}

\usepackage{booktabs}
\usepackage{graphicx}
\usepackage{subcaption}
\usepackage{multirow}
\usepackage[
  top=4.3cm,
  right=4.8cm,
  bottom=4.3cm,
  left=4.8cm
]{geometry}
\usepackage{pifont}
\usepackage{url}
\usepackage{hyperref}
\hypersetup{hidelinks}
\newcommand{\cmark}{\ding{51}}%
\newcommand{\xmark}{\ding{55}}%

\journal{Pattern Recognition}

\begin{document}

\begin{frontmatter}

\title{Multi-source Transfer Learning of Time Series with a Shapelet-based Distance Measure}

\author[1]{Jiseok Lee\corref{cor1}}
\ead{jiseok.lee@human.ait.kyushu-u.ac.jp}
\author[2]{Brian Kenji Iwana}
\ead{iwana@ait.kyushu-u.ac.jp}
\cortext[cor1]{Corresponding Author}
\affiliation[1]{organization={Graduate School of Information Science and Electrical Engineering, Kyushu University}, 
addressline={744 Motooka, Nishi Ward}, 
city={Fukuoka},
postcode={819-0395}, 
state={Fukuoka},
country={Japan}}
\affiliation[2]{organization={Faculty of Information Science and Electrical Engineering, Kyushu University}, 
addressline={744 Motooka, Nishi Ward}, 
city={Fukuoka},
postcode={819-0395}, 
state={Fukuoka},
country={Japan}}

\begin{abstract}
Transfer learning is an effective technique for addressing data scarcity in deep learning for time series classification, but its success depends on the selection of source datasets. Conventional transferability estimation methods are often computationally expensive, as they require fully pre-training a model on each potential source dataset to assess its suitability. This paper introduces a novel, training-free source selection method named Shapelet Matching. Our approach first identifies discriminative shapelets from the target and potential source datasets. Then, Shapelet Matching quantifies dataset similarity by comparing the extracted sets of shapelets. To mitigate the risk of negative transfer from selecting an unsuitable single source, we introduce a multi-source transfer learning method. We select several source datasets based on their shapelet-based similarity scores, combine them into a single multi-source dataset, and use this aggregated dataset for pre-training. The model is then fine-tuned on the target task. We evaluated our method on 128 datasets from the UCR Archive using both temporal CNN and Transformer architectures. The empirical results demonstrate that our multi-source pre-training reduces the risk of negative transfer on average. Shapelet Matching achieves the strongest performance for the CNN backbone and remains competitive for patch-based Transformer architectures, while avoiding the cost of pre-training a separate model for every candidate source.

\end{abstract}

\begin{keyword}

Transfer learning \sep Time series classification \sep Transferability estimation

\end{keyword}

\end{frontmatter}

\section{Introduction}
Deep neural networks have become a dominant and effective methodology for time series recognition.
For instance, temporal convolutional neural networks (CNN)~\cite{Lecun_1998} have demonstrated considerable success across various time series domains.
A primary limitation of these models, however, is their reliance on large training datasets~\cite{Ismail_Fawaz_2018,iwana2021an}.
This dependency presents a significant challenge, as most time series data is unlabeled, making it costly to acquire sufficient labeled examples for supervised learning.

To address the challenge of limited labeled data, several techniques such as transfer learning, self-supervised learning, and data augmentation have proven effective.
Among these, transfer learning is an effective method for initializing the weights of neural networks~\cite{Bozinovski_2020}.
In this study, we adopt a standard pre-training and fine-tuning paradigm.
A model is first trained on a large-scale source dataset and subsequently adapted to a smaller target dataset.
This process can significantly reduce the data requirements for the target task by enabling the model to learn general-purpose feature representations during pre-training~\cite{zhuang2020comprehensive}.
Recent time series foundation models, including Timer~\cite{liu2024timer}, MOMENT~\cite{goswami2024moment}, and UniTS~\cite{gao2024units}, further highlight the value of large-scale pre-training. However, many practical supervised time-series classification settings still lack access to large external corpora or released foundation-model checkpoints.
In such cases, users are often faced with a finite pool of heterogeneous labeled datasets and must decide which of them should be used as sources for pre-training.

However, the usefulness of pre-training often depends on the domain gap or similarity between the source and target datasets.
Such a gap may arise from differences in temporal scale, sampling characteristics, noise patterns, or the location and granularity of class-discriminative patterns.
Consequently, source selection is critical because pre-training on mismatched sources can bias the feature extractor toward non-transferable patterns and induce negative transfer.
A common way to address this problem is to estimate transferability using a model that has already been pre-trained on each candidate source.
However, for time series datasets, this model-based strategy can be costly because every candidate source must be pre-trained for each backbone architecture before it can be evaluated.
It is also less transparent than pattern-based comparison, since the selected source is justified by model outputs or internal features rather than by explicit temporal subsequences.

To address the scarcity of large, suitable datasets for pre-training, we propose a multi-source transfer learning approach.
This method involves consolidating multiple smaller datasets into a single, larger source dataset for pre-training.
Multi-source pre-training can reduce dependence on a single source, but it also makes source selection more important because mismatched sources can still introduce negative transfer.
To mitigate this risk, we propose using shapelet-based similarity as a metric to assess the transferability between potential source datasets and the target task, thereby guiding the selection process.
Our method is based on the assumption that transferable local discriminative subsequences exist across at least a subset of source and target datasets. When two datasets share such shapelets, pre-training on the source is more likely to induce representations that remain useful after fine-tuning on the target.

\begin{figure}[t]
    \centering
    \includegraphics[width=\linewidth]{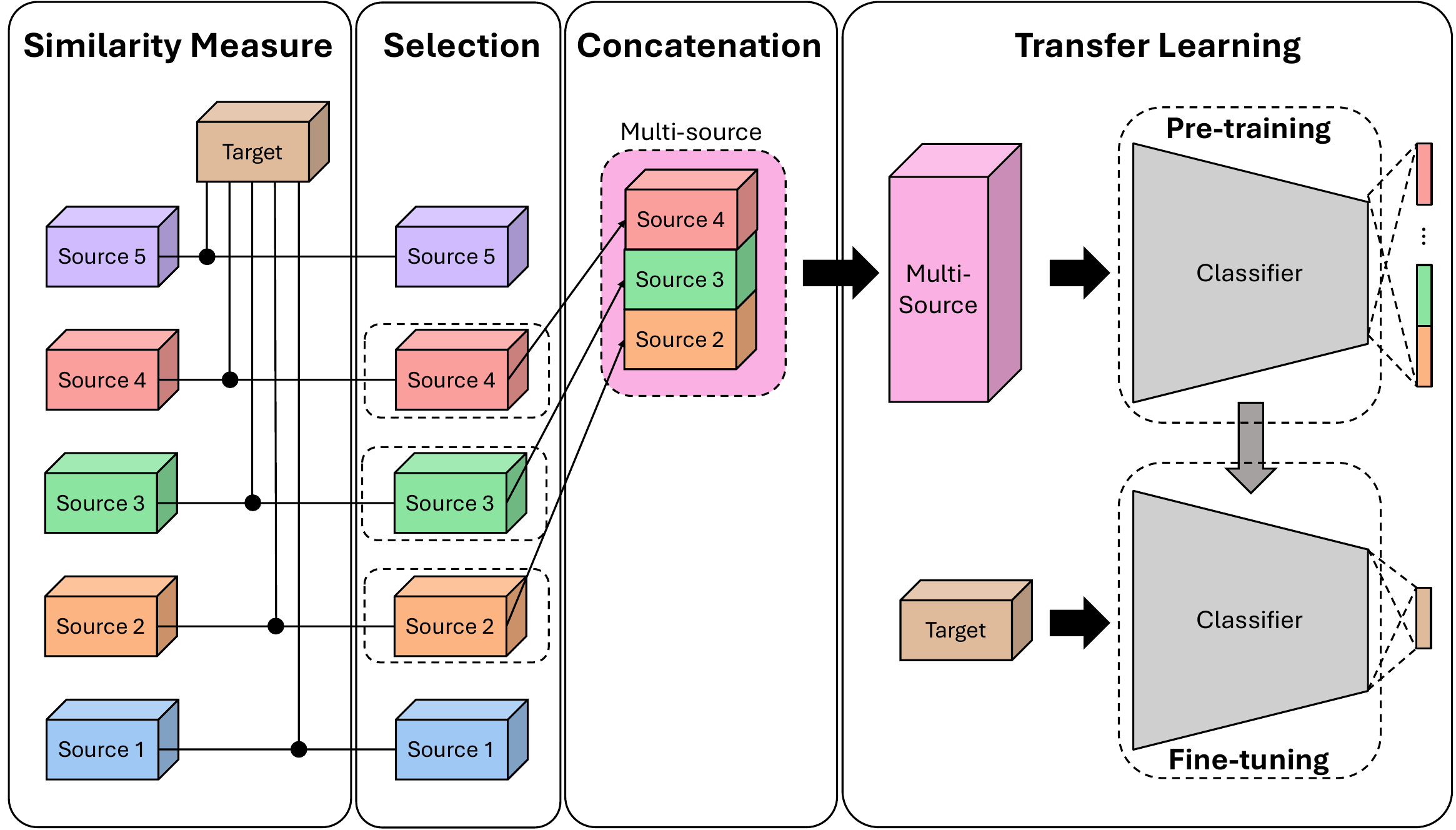}
    \caption{End-to-end overview of the proposed multi-source transfer learning framework. In the \textit{Similarity Measure} stage, class-discriminative shapelets are discovered from the target dataset and each candidate source dataset using the matrix profile-based procedure, and each source is assigned a Shapelet Matching distance to the target. In the \textit{Selection} stage, sources are ranked by this distance and the top-$K$ sources (dashed boxes) are retained. In the \textit{Concatenation} stage, the selected sources are resized to the target temporal length, their label spaces are remapped into a unified multi-source label set, and the datasets are balanced so that each source contributes the same number of samples. In the \textit{Transfer Learning} stage, a backbone network is pre-trained on the aggregated multi-source dataset and then fine-tuned on the target dataset with a target-specific classifier head.}

    \label{fig:abst}
\end{figure}

This paper proposes a novel source selection method for multi-source transfer learning based on a shapelet-based distance measure, which serves as an alternative to conventional transferability estimation.
The proposed multi-source transfer learning framework, shown in Fig.~\ref{fig:abst}, comprises four steps.
First, class-discriminative shapelets are identified from each dataset using a matrix profile-based discovery algorithm~\cite{Yeh_2016}.
Second, each candidate source is scored and ranked according to its shapelet-based distance to the target dataset.
Third, the selected sources are resized, balanced, relabeled, and concatenated into a unified multi-source dataset.
Finally, the backbone is pre-trained on the aggregated source dataset and fine-tuned on the target task.
The underlying principle of this approach is that datasets sharing discriminative subsequences are presumed to possess similar feature distributions, thereby enabling effective knowledge transfer.
Because the proposed criterion is computed directly from source and target datasets, it does not require pre-training a model for each candidate source. 
The resulting source rankings are therefore reusable across backbone  architectures and offer an explicit pattern-level rationale through the matched shapelets.
This combination targets a practical gap between conventional transferability estimation and large-scale foundation-model pre-training: it keeps source selection lightweight while retaining an explicit pattern-level explanation of why a source is selected.

The contributions of this paper are as follows:
\begin{itemize}
    \item We propose a new shapelet-based similarity measure called Shapelet Matching. This method focuses on dense matching of subsequences extracted from the target dataset and the candidate source datasets.
    \item We integrate this criterion into a multi-source transfer learning framework that aggregates the top-ranked source datasets for pre-training and then fine-tunes the backbone on the target task.
    \item We evaluate the method on all 128 univariate datasets from the 2018 UCR Time Series Archive (UCR Archive)~\cite{UCRArchive2018}, using CNN, Vision Transformer (ViT)~\cite{dosovitskiy2020image}, and PatchTST~\cite{nie2022time} backbones.
    \item We report statistical significance tests, computational cost comparisons, and architecture-dependent behavior to clarify both the strengths and limitations of the proposed method. We note that the method’s effectiveness varies by backbone architecture, which is further discussed in Section~\ref{sec:MD}.
    \item We provide the implementation code for multi-source transfer learning with shapelet similarity-based source selection on GitHub\footnote{\url{https://github.com/uchidalab/shapelet-matching}}.
\end{itemize}

\section{Related Work}
\label{chap:bg}

\subsection{Transfer Learning for Time Series}
\label{chap:tlfts}
Transfer learning has long been recognized as an effective strategy when labeled data in the target domain are limited, and Zhuang et al.~\cite{zhuang2020comprehensive} provide a broad survey of this literature. For time series classification (TSC), however, the literature remains substantially smaller than in computer vision. Fawaz et al.~\cite{Ismail_Fawaz_2018} provided one of the earliest large-scale studies of transfer learning for TSC, showing on the UCR Archive that pre-training a CNN on a source dataset can improve target accuracy, while also revealing the risk of negative transfer when the source and target are poorly matched. Weber et al.~\cite{Weber_2021} later systematized the broader literature on transfer learning with time series data and highlighted that TSC-specific studies with temporal neural networks are still relatively limited. More application-oriented work has explored transfer in specific settings: Clark and Doyle~\cite{Clark_2022} studied transferability in cyber-physical health systems, whereas Gikunda and Jouandeau~\cite{Gikunda_2021} considered homogeneous transfer active learning for time series classification. Taken together, these studies suggest that transfer learning is promising for TSC, but that its success depends strongly on source--target compatibility.

A more recent line of work studies large pre-trained models for time series.
Zhou et al.~\cite{zhou2023one} showed that a pre-trained language model can be adapted to multiple time series tasks within a unified framework.
Timer~\cite{liu2024timer}, MOMENT~\cite{goswami2024moment}, and UniTS~\cite{gao2024units} further push this direction by learning general-purpose representations from large and heterogeneous time series data.
These models are powerful when large pre-training data and substantial computational resources are available.
However, this setting differs from the practical transfer-learning scenario studied in this paper, where users have limited computational resources and need task-specific portable models.
Our Shapelet Matching method is therefore complementary to time-series foundation models rather than a replacement for them. 
Our method provides a lightweight, training-free source-selection criterion whose matched subsequences offer potential interpretability when full-scale foundation-model pre-training is unavailable, unnecessary, or computationally prohibitive.

Recent shape-aware time-series classification research has also advanced efficiency and interpretability.
SoftShape learns sparsified shape representations for efficient classification~\cite{liu2025learning}, whereas UniShape pre-trains a shape-aware foundation model to capture transferable patterns across domains~\cite{liu2026unishape}.
In contrast to these approaches, which learn shape-aware representations within the classifier, Shapelet Matching uses discovered shapelets as a training-free criterion for ranking candidate source datasets before downstream model training.

\subsection{Multi-Source Transfer Learning}
Research on multi-source transfer learning, which leverages multiple source datasets for pre-training, is relatively less explored.
The existing work in this area can be broadly categorized.
One category includes boosting-based methods, such as the application of TrAdaBoost~\cite{Dai_2007} by Yao et al.~\cite{Yao_2010} and the SharedBoost algorithm proposed by Huang et al.~\cite{Huang_2012}.
Another category focuses on developing novel learning frameworks or weighting methods; for example, Multi-transfer~\cite{Tan_2013} combines multi-view with multi-source transfer learning, while Song et al.~\cite{Song_2017} introduced a method to weight source domains based on conditional probability differences.

Multi-source transfer learning has also been specifically applied to time series classification. 
Some approaches focus on selecting the most relevant sources for a target domain.
In the context of electroencephalogram (EEG) data, for example, Li et al.~\cite{Li_2019} trained a separate model for each source and selected the top-performing one, while Ren et al.~\cite{Ren_2022} employed a preliminary classifier for the same purpose.
Other methodologies aim to combine information from all sources.
For sensor modality classification, Li et al.~\cite{Li_2020} trained a neural network to predict the source of time series segments, thereby learning an aligned representation.

Furthermore, this research presents several significant extensions of our preliminary study~\cite{lee2024model}. 
In this work, we propose the Shapelet Matching method, a substantial generalization of our previously introduced Minimum Shapelet method, by densely matching mutually preferable subsequences rather than matching only single pairs.
Additionally, we have evaluated our methods on CNNs, ViT~\cite{dosovitskiy2020image}, and PatchTST~\cite{nie2022time}, and based on the ViT results, we provide a more precise analysis of architecture dependency.
Finally, we evaluate multi-source training with conventional transferability measures, which was not examined in our prior work.

\section{Transferability Measures}
\subsection{Problem Setting}
\sloppy Assume a neural network $\theta = (\omega, h)$, where $\omega$ is a feature extractor and $h$ is a linear classifier (head). 
Consider a transfer learning scenario where a neural network $\theta$ is first pre-trained on a source dataset $\mathcal{S}$:
\begin{equation}
    \mathcal{S} = \bigl\{\bigl(\mathbf{s}_1, z_1\bigr), \dots, \bigl(\mathbf{s}_m, z_m\bigr),\dots,  \bigl(\mathbf{s}_M, z_M\bigr)\bigr\}.
\end{equation}
Let $\mathbf{S} = \{\mathbf{s}_1, \dots, \mathbf{s}_m, \dots, \mathbf{s}_M\}$ be the set of source samples, and
$\mathbf{Z} = \{z_1, \dots, z_m, \dots z_M\}$ be the corresponding source labels, with each label $z_m \in \{1, \dots,z, \dots, C_S\}$ where $C_S$ represents the number of source classes. 
The network $\theta$ is then fine-tuned on a {target dataset} $\mathcal{T}$:
\begin{equation}
\mathcal{T} = \bigl\{\bigl(\mathbf{x}_1, y_1\bigr), \dots, \bigl(\mathbf{x}_n, y_n\bigr), \dots, \bigl(\mathbf{x}_N, y_N\bigr)\bigr\}.
\end{equation}
Let $\mathbf{X} = \{\mathbf{x}_1, \dots, \mathbf{x}_n, \dots, \mathbf{x}_N\}$ be the set of target samples, and
let $\mathbf{Y} = \{y_1, \dots, y_n, \dots, y_N\}$ be the corresponding target labels, where $y_n \in \{1, \dots, C_T\}$ and $C_T$ denotes the number of target classes.
Unlike the domain adaptation problem setting, there is no assumption that the datasets are similar tasks, that source label $z_{m}$ and target label $y_n$ are related, or that there exists a hypothesis (or model) that is suitable for both datasets~\cite{zhang2022survey}.

Our setting is related to, but distinct from, standard domain adaptation and domain generalization. 
Domain adaptation typically assumes the same or closely related prediction task across domains and aims to reduce distribution shift~\cite{ragab2022self}, whereas domain generalization learns from multiple source domains to generalize to unseen domains~\cite{deng2024domain}.
By contrast, our setting allows heterogeneous source datasets with dataset-specific label spaces and potentially different task semantics.
The objective is not domain alignment or domain-invariant prediction, but target-guided source selection: identifying source datasets that provide useful pre-training signals before fine-tuning on the target task.

\subsection{Transferability Estimation}
Given a source-trained model $\theta=(\omega, h)$, a transferability estimator predicts how well the model is expected to perform after fine-tuning on $\mathcal{T}$, without explicitly fine-tuning the model. The estimator computes a score from the target labels $\mathcal{Y}$ together with either the source-model predictions $\theta(\mathbf{X})$ or the source-trained features $\omega(\mathbf{X})$. The source and target label spaces need not correspond; the source predictions are used only to construct the transferability score.

\subsubsection{H-score}
H-score~\cite{Bao_2019} is a transferability estimator computed from the features extracted by $\omega$ and the target labels $\mathbf{Y}$, or:
\begin{equation}
    \text{H-score} (\mathcal{S}, \mathcal{T})=\mathrm{Tr}\left(\mathrm{cov}(\omega ( \mathbf{X}))^{-1} \mathrm{cov}\left(\mathbb{E}[\omega(\mathbf{X}) \mid \mathbf{Y}]\right)\right),
\end{equation}
where $\mathrm{Tr}(A) = \sum_{i} (A)_{ii}$, and $\mathbb{E}[\omega(\mathbf{X}) \mid \mathbf{Y}]$ is the mean of a set of features from each label.

\subsubsection{Log Expected Empirical Prediction (LEEP)}
LEEP~\cite{pmlr-v119-nguyen20b} attempts to predict the transferability for $\mathcal{T}$ using $\theta$ pre-trained on $\mathcal{S}$. 
LEEP is then calculated by:
\begin{equation}
    \mathrm{LEEP}(\mathcal{S}, \mathcal{T}) = \frac{1}{N}\sum^{N}_{n=1}\mathrm{log}\left(\sum_{z=1}^{C_S}\hat{P}(y_n | z)\theta(\mathbf{x}_n)[z]\right),
\end{equation}
where $\theta(\mathbf{x}_n)[z]$ is the probability assigned by the source-trained model to source class $z$ for the input $\mathbf{x}_n$, and $\hat{P} \bigl(y | z\bigr)$ is an empirical conditional distribution estimating how likely target label $y$ is, given that the model’s prediction lies in source class $z$. Specifically:
\begin{equation}
    \hat{P}(y \mid z)=\frac{\hat{P}(y,z)}{\hat{P}(z)}=\frac{\sum_{n: y_n=y}\theta(\mathbf{x}_n)[z]}{\sum^N_{n=1}\theta(\mathbf{x}_n)[z]}.
\end{equation}

\subsubsection{Log Maximum Evidence (LogME)}
LogME~\cite{pmlr-v139-you21b} is a transferability estimation that assesses how well pre-trained $\omega$ can explain $\mathbf{Y}$.
Specifically, the key quantity of LogME is the marginal log-likelihood (evidence):
\begin{equation}
\mathcal{L}(\alpha, \beta) = \log P(\mathbf{Y}  |  \omega(\mathbf{X}), \alpha, \beta).
\end{equation}
LogME finds $\alpha^*$ and $\beta^*$ that maximize the evidence, $\mathcal{L}(\alpha, \beta)$.
LogME is then calculated by normalizing the maximum evidence by $N$:
\begin{equation}
    \mathrm{LogME}(\mathcal{S}, \mathcal{T}) = \frac{\mathcal{L}(\alpha^*, \beta^*)}{N},
\end{equation}
where
\begin{equation}
     (\alpha^*, \beta^*) = \mathrm{argmax}_{\alpha,\beta}\mathcal{L}(\alpha, \beta).
\end{equation}

\subsubsection{Negative Conditional Entropy (NCE)}
NCE~\cite{Tran_2019} estimates transferability by measuring the uncertainty of target labels given source predictions. A lower conditional entropy suggests that source labels provide more information about target labels, implying better transferability.

NCE is computed as:
\begin{equation} \mathrm{NCE}(\mathcal{S}, \mathcal{T}) = \sum_{y=1}^{C_T} \sum_{z=1}^{C_S} \hat{P}(y, z) \log \frac{\hat{P}(y, z)}{\hat{P}(z)}, \end{equation}
where $\hat{P}(y,z)$ is the empirical joint probability of target label $y$ and predicted source label $z$, and $\hat{P}(z)=\sum_{y=1}^{C_T}\hat{P}(y,z)$.

\subsubsection{TransRate}
Similar to LogME, TransRate~\cite{pmlr-v162-huang22d} measures how well the features $\omega(\mathbf{X})$ are separated into class-specific clusters.
Formally, TransRate is defined as:
\begin{equation}
    \mathrm{TransRate}(\mathcal{S}, \mathcal{T}) = R(\omega(\mathbf{X}), \epsilon) - R(\omega(\mathbf{X}), \epsilon|\mathbf{Y}),
\end{equation}
where $R(\omega(\mathbf{X}), \epsilon)$ is the coding rate, measuring how many bits are required to represent $\omega(\mathbf{X})$ under a distortion $\epsilon$:
\begin{equation}
    R(\omega(\mathbf{X}), \epsilon) = \frac{1}{2}\log \mid \mathbf{I} + \frac{1}{N\epsilon}\omega(\mathbf{X})^T\omega(\mathbf{X})\mid,
\end{equation}
and $R(\omega(\mathbf{X}), \epsilon \mid \mathbf{Y})$ is the conditional coding rate computed by averaging coding rates over each class.
Because features that are well-separated by class are easier to compress, the conditional coding rate tends to be lower when classes form distinct clusters.

\subsubsection{Transferability Measurement with Intra-class feature Variance (TMI)}
TMI~\cite{xu2023fast} focuses on generalization ability by calculating intra-class variance, whereas other measures focus on the clustering ability of pre-trained models, or:
\begin{equation}
    \mathrm{TMI} (\mathcal{S}, \mathcal{T}) = H(\omega(\mathbf{X})|\mathbf{Y}) = \sum_{y=1}^{C_T} \frac{N_y}{N}H(\omega(\mathbf{X}_y)),
\end{equation}
where $H(\omega(\mathbf{X})|\mathbf{Y})$ is the conditional entropy of the target feature $\omega(\mathbf{X})$ on the model pre-trained with $\mathcal{S}$ given the target labels $Y$, $\mathbf{X}_y$ is the subset of $\mathbf{X}$ belonging to the label $y$, and $N_y$ is the number of instances in class $y$.

\subsubsection{Dataset Similarity Measure for Source Selection}
Similar to the transferability estimation, dataset similarity measures are often used for source selection.
The previous method, transferability estimation, is only useful when the pre-trained $\theta$ is available.
However, unlike image recognition, there are fewer standard pre-trained weights available for time series recognition.
Therefore, to find a suitable source with those transferability estimations for time series recognition, pre-training all available datasets is necessary, which incurs substantial computational cost.

By contrast, dataset distance measures only require information from the datasets. 
Fawaz et al.~\cite{Ismail_Fawaz_2018} demonstrated that Dynamic Time Warping (DTW)~\cite{SAKOE_1990} can be used to compare prototypes of each time-series class in the target and source datasets.
The prototypes are obtained by computing the average time series of each class found by DTW Barycenter Averaging (DBA)~\cite{Petitjean_2011}. 
They defined the dataset distance as the distance between the most similar pairs of prototypes gained by DBA from each dataset~\cite{Ismail_Fawaz_2018}.
Using the dataset distance measure for source selection helped to find an appropriate source dataset for the target task.
A benefit of this method and our proposed method is that these methods do not require a trained model to predict transferability.

\section{Multi-Source Transfer Learning}
\label{sec:multi}

We propose a multi-source transfer learning framework that aggregates the top-$K$ source datasets selected by the shapelet-based similarity measure into a single unified pre-training dataset, denoted as $\mathcal{S}_{\mathrm{Multi}}$. Let $\mathcal{S}_{\mathrm{sel}}=\{\mathcal{S}_{\pi_1},\dots,\mathcal{S}_{\pi_K}\}$ denote the selected source datasets ranked by their similarity to the target dataset $\mathcal{T}$. An overview of the framework is shown in Fig.~\ref{fig:abst}, and the complete end-to-end procedure is summarized in Algorithm~\ref{alg:overall_pipeline}.

\begin{algorithm}[htbp]
\caption{End-to-end multi-source transfer learning with Shapelet Matching}
\label{alg:overall_pipeline}
\begin{algorithmic}[1]
\Require Target dataset $T$, candidate source datasets $\mathcal{S}=\{S_1,\dots,S_I\}$, number of selected sources $K$, shapelet length $\ell_s$, number of shapelets per class $Q$, backbone $f_\theta$, pre-training iterations $N_{\mathrm{pre}}$, fine-tuning iterations $N_{\mathrm{ft}}$
\Ensure Fine-tuned target model $g_\psi \circ f_\theta$
\State $P^{(T)} \gets \textsc{DiscoverShapeletsMP}(T,\ell_s,Q)$
\For {$i \gets 1$ \textbf{to} $I$}
    \State $P^{(S_i)} \gets \textsc{DiscoverShapeletsMP}(S_i,\ell_s,Q)$
    \State $d_i \gets \textsc{ShapeletMatchingDistance}(P^{(S_i)}, P^{(T)})$
\EndFor
\State $\pi \gets \textsc{ArgsortAscending}(d_1,\dots,d_I)$
\State $\mathcal{S}_K \gets \{S_{\pi_1},\dots,S_{\pi_K}\}$
\State $\ell_T \gets \textsc{TemporalLength}(T)$
\State $\hat{\mathcal{S}}_K \gets \varnothing$
\For {$S \in \mathcal{S}_K$}
    \State $\hat{\mathcal{S}}_K \gets \hat{\mathcal{S}}_K \cup \{\textsc{ResizeToLength}(S,\ell_T)\}$
\EndFor
\State $M^\star \gets \max_{\hat{S} \in \hat{\mathcal{S}}_K} |\hat{S}|$
\State $S_{\mathrm{Multi}} \gets \varnothing$
\State $o \leftarrow 0$
\For {$\hat{S} \in \hat{\mathcal{S}}_K$}
    \State $\tilde{S} \gets \textsc{OversamplePreservingClassRatio}(\hat{S}, M^\star)$
    \State $(\tilde{X},\tilde{Z},o) \gets \textsc{RelabelWithOffset}(\tilde{S}, o)$
    \State $S_{\mathrm{Multi}} \gets S_{\mathrm{Multi}} \cup (\tilde{X},\tilde{Z})$
\EndFor
\State $C_{\mathrm{Multi}} \gets |\{z \mid (\mathbf{x},z)\in S_{\mathrm{Multi}}\}|$
\State $(f_\theta,h_\phi) \gets \textsc{InitializeModel}(C_{\mathrm{Multi}})$
\State $(f_\theta,h_\phi) \gets \textsc{TrainForIterations}(f_\theta,h_\phi,S_{\mathrm{Multi}},N_{\mathrm{pre}})$
\State $C_T \gets |\{y \mid (\mathbf{x},y)\in T\}|$
\State $g_\psi \gets \textsc{InitializeTargetHead}(C_T)$
\State $(f_\theta,g_\psi) \gets \textsc{TrainForIterations}(f_\theta,g_\psi,T,N_{\mathrm{ft}})$
\State \Return {$g_\psi \circ f_\theta$}
\end{algorithmic}
\end{algorithm}

After source selection, the retained datasets undergo several pre-processing steps to enable multi-source pre-training.
First, all selected source datasets in $\mathcal{S}_{\mathrm{sel}}$ are resampled to match the temporal length of the target dataset $\mathcal{T}$.
This operation ensures that the neural network input dimensionality remains consistent during multi-source pretraining.
If the target sequence length exceeds that of a source dataset, interpolation-based upsampling is applied to expand the source sequences to the target length.
However, since the primary objective of pre-training is to obtain effective initial weights for the target domain, the potential loss of source-specific characteristics is acceptable.
Second, the aggregated dataset $\mathcal{S}_{\mathrm{Multi}}$ is balanced to ensure that each constituent source dataset $\mathcal{S}_i$ contributes an equal number of time series samples.
This is achieved via oversampling while maintaining the original class distribution within each source.
Finally, to ensure a fair comparison across experiments with varying sizes of $\mathcal{S}_{\mathrm{Multi}}$, the model is trained for a fixed number of iterations rather than for a fixed number of epochs.

After preprocessing, the selected source datasets in $\mathcal{S}_{\mathrm{sel}}$ are merged by modifying the label space.
Specifically, the corresponding label sets from each dataset, ${\mathbf{Z}}_{\pi 1},\dots,{\mathbf{Z}}_{\pi k},\dots,{\mathbf{Z}}_{\pi K}$, are concatenated to form a single, unified label set $\mathbf{Z}_\mathrm{Multi}$.
Consequently, the output layer of the network is extended to accommodate this expanded set of classes.
The network is then pre-trained on the aggregated dataset $\mathcal{S}_\mathrm{Multi}$ using the unified labels $\mathbf{Z}_\mathrm{Multi}$.
We emphasize that this aggregation is used to learn transferable feature extractors rather than to merge class semantics across domains. Each source label remains dataset-specific after relabeling, and pre-training exposes the network to a broader collection of discriminative temporal patterns. The benefit of aggregation is therefore not semantic equivalence across sources, but improved initialization from diverse yet target-relevant sources.
This strategy enables the pre-training of a model on a substantially larger and more diverse dataset than would be possible with a single source.

After multi-source pre-training, the multi-source classifier head is replaced with a target-specific classifier head, and the transferred backbone is fine-tuned on the target task following the standard transfer learning procedure.
Although the source-selection score can be computed independently of the downstream backbone, its empirical effectiveness may still depend on the model architecture.
This architecture-dependent behavior is analyzed in Section~\ref{sec:MD} and discussed as a limitation in Section~\ref{sec:lim}.

\section{Shapelet Similarity-based Source Selection}
It is essential to select a number of source datasets for the proposed multi-source transfer learning.
However, selecting proper source datasets is not a straightforward task.
Thus, for source selection, we propose to use a novel shapelet-based similarity measure instead of using conventional transferability measures.

\subsection{Shapelet}
A shapelet is a subsequence of time series data that is highly discriminative for a given class~\cite{Ye_2009}.
Shapelets imply class-representative patterns or features of each class.

\subsection{Shapelet Discovery}
From a time series, a shapelet can be any subsequence depending on its definition.
Thus, sometimes finding the most discriminative shapelet can be computationally expensive.
In this research, we used matrix profile~\cite{Yeh_2016} to find discriminative shapelets for efficient shapelet discovery.

Matrix profile represents a time series as distances between subsequences and their nearest neighbor. 
Matrix profile $\mathbf{p}$ of given time series $\mathbf{t}$ and its list of all subsequences $\mathcal{A}$ is a sequence of the distances between every subsequence $\mathcal{A}_r$ to its nearest neighbor, or:
\begin{equation}
\label{eq:differences}
    \mathbf{p} =  \mid  \mid \mathcal{A}_{1} - \mathcal{E}_{1} \mid  \mid , \dots,  \mid  \mid \mathcal{A}_{r} - \mathcal{E}_{r} \mid  \mid , \dots,  \mid  \mid \mathcal{A}_{R} - \mathcal{E}_{R} \mid  \mid.
\end{equation}
Here, $\mathcal{E}_r$ denotes the subsequence of $\mathcal{A}$ closest to the corresponding subsequence $\mathcal{A}_r$, and $\|\cdot\|$ denotes the Euclidean norm.
By computing the matrix profile $\mathbf{p}$ from the time series $\mathbf{t}$, motifs and discords can be efficiently identified.

To apply matrix profile for shapelet discovery, we introduce several modifications. Given a dataset $\mathcal{S}$, we first concatenate all time series belonging to each class $c$ into a single class-specific time series $\mathbf{t}^{(c)}$. For instance, if there are two classes (e.g., class 1 and class 2), two concatenated series $\mathbf{t}^{(1)}$ and $\mathbf{t}^{(2)}$ are created.
Next, instead of calculating the matrix profile solely using nearest neighbors within the same class (as in \eqref{eq:differences}), we calculate matrix profiles across all class combinations: $\mathbf{p}^{(1,1)}$, $\mathbf{p}^{(1,2)}$, $\mathbf{p}^{(2,2)}$, and $\mathbf{p}^{(2,1)}$. Subsequently, the largest values in the difference profiles $\mathbf{p}^{(1,2)}-\mathbf{p}^{(1,1)}$ and $\mathbf{p}^{(2,1)}-\mathbf{p}^{(2,2)}$ correspond to the most representative shapelet candidates for class 1 and class 2, respectively.
As this procedure inherently supports binary classification, we generalize the approach for multi-class scenarios by employing a one-versus-all strategy to identify distinctive shapelets for each class separately.
In implementation, time series from the same class are concatenated with boundary separators before matrix profile computation.
Windows yielding invalid matrix-profile values near sequence boundaries are assigned invalid scores and deprioritized during candidate ranking, which prevents candidates from spanning two original series.

\subsection{Source Selection based on Shapelet Similarity}
\begin{figure}[htpb]
\centering
\includegraphics[width=\columnwidth]{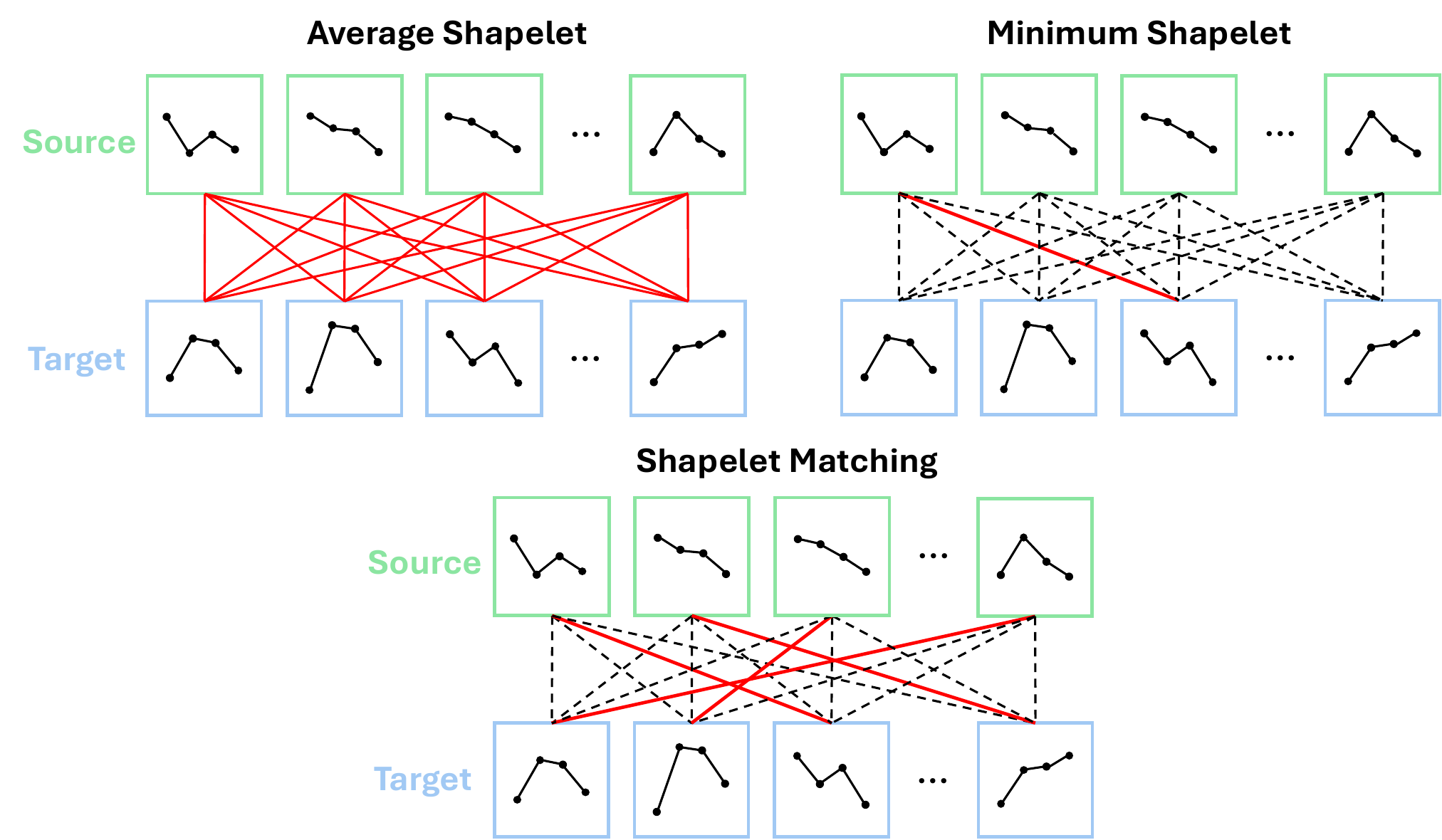}
\caption{Comparison of the three shapelet-based dataset-similarity measures. 
Green boxes denote shapelets extracted from a source dataset, and blue boxes denote shapelets extracted from the target dataset. 
Red connections indicate the source--target shapelet pairs whose Euclidean distances are averaged to compute the dataset distance, whereas dashed black connections indicate candidate pairs that are not used in the final aggregation. 
Average Shapelet averages all possible pairs, Minimum Shapelet uses only the closest pair, and Shapelet Matching greedily selects multiple non-overlapping closest pairs.}
\label{fig:shapelet_method}
\end{figure}

We propose the shapelet-based similarity measures using Shapelet Matching for source selection, each of which compares the class-representative shapelets $\mathcal{P}^{(\mathcal{S})}$ and $\mathcal{P}^{(\mathcal{T})}$ from the source and target datasets, respectively.
Notably, the other two shapelet-based source selection schemes, Minimum Shapelet and Average Shapelet, are from the preliminary conference paper of this work~\cite{lee2024model}.
Average Shapelet summarizes similarity across all source--target
shapelet pairs, Minimum Shapelet emphasizes the closest local match,
and Shapelet Matching balances these extremes by aggregating
multiple greedy matches.
Figure~\ref{fig:shapelet_method} illustrates how each scheme handles pairing $\mathcal{P}^{(\mathcal{S})}$ and $\mathcal{P}^{(\mathcal{T})}$.
All three schemes rely on pairwise Euclidean distances but adopt distinct strategies to match and average those distances, leading to different similarity scores.

\textbf{Average Shapelet} measures the shapelet similarity from the mean distance of all possible pairs of $\mathcal{P}^{(\mathcal{S})}_i$ and $\mathcal{P}^{(\mathcal{T})}_j$, or:
\begin{equation}
    D_\mathrm{as} = \mathop{\mathbb{E}}_{i,j} (\mid  \mid \mathcal{P}^{(\mathcal{S})}_i - \mathcal{P}^{(\mathcal{T})}_j \mid  \mid ),
\end{equation}
where $\mathcal{P}^{(\mathcal{S})}_i$ and $\mathcal{P}^{(\mathcal{T})}_j$ are the $i$-th and $j$-th shapelet of $\mathcal{P}^{(\mathcal{S})}$ and $\mathcal{P}^{(\mathcal{T})}$, respectively.

\textbf{Minimum Shapelet} is inspired by the DBA-DTW~\cite{Ismail_Fawaz_2018} that measures DBA-based similarity with its most similar pair.
This measures the shapelet similarity by measuring the distance of the most similar pair between source and target shapelets, $\mathcal{P}^{(\mathcal{S})}_i$ and $\mathcal{P}^{(\mathcal{T})}_j$, or:
\begin{equation}
D_\mathrm{ms}=\min_{i,j}  \mid  \mid \mathcal{P}^{(\mathcal{S})}_i - \mathcal{P}^{(\mathcal{T})}_j \mid  \mid .
\end{equation}

\textbf{Shapelet Matching} is defined using a greedy matching approach.
The primary motivation for Shapelet Matching is to balance the sparse matching of Minimum Shapelet and the dense matching of Average Shapelet.
Shapelet Matching iteratively selects the closest pairs based on a pre-computed distance matrix between two sets of shapelets.
Given a set of source shapelets and a set of target shapelets, the algorithm identifies and matches pairs based on their Euclidean distance between each pair of shapelets, ensuring that each target shapelet is matched exactly once, while each entry in the expanded source-shapelet list is used at most once.

\begin{algorithm}[h!]
\caption{Shapelet Matching Distance}
\label{alg:sm_shapelet}
\begin{algorithmic}[1]
\Require $\mathcal{P}^{(\mathcal{S})}=\{s_1,\dots,s_m\}$ (source shapelets), 
         $\mathcal{P}^{(\mathcal{T})}=\{t_1,\dots,t_n\}$ (target shapelets)
\Ensure Matched triples $M$ and mean pairwise distance $\bar d$

\State $m \gets |\mathcal{P}^{(\mathcal{S})}|$, $n \gets |\mathcal{P}^{(\mathcal{T})}|$

\If{$m < n$}
    \State $r \gets \left\lceil n/m \right\rceil$
    \State $\widetilde{\mathcal{P}}^{(\mathcal{S})} 
           \gets \operatorname{Repeat}(\mathcal{P}^{(\mathcal{S})}, r)$
           \Comment{keep duplicated shapelets as distinct entries}
\Else
    \State $\widetilde{\mathcal{P}}^{(\mathcal{S})} 
           \gets \mathcal{P}^{(\mathcal{S})}$
\EndIf

\State $\widetilde{\mathcal{P}}^{(\mathcal{S})}
       =\{\tilde{s}_1,\dots,\tilde{s}_{m'}\}$, 
       where $m' = |\widetilde{\mathcal{P}}^{(\mathcal{S})}|$
\State $I \gets \{1,\dots,m'\}$ \Comment{active source-shapelet indices}
\State $J \gets \{1,\dots,n\}$ \Comment{active target-shapelet indices}
\State $M \gets [\,]$ \Comment{matched triples}
\State $D \gets 
       \bigl[\lVert \tilde{s}_i - t_j \rVert_2\bigr]_
       {i=1,\dots,m',\;j=1,\dots,n}$
       \Comment{pre-computed distance matrix}

\While{$J \neq \emptyset$}
    \State $(i^\star,j^\star) 
           \gets \arg\min_{i\in I,\;j\in J} D_{i,j}$
    \State $\operatorname{append}
           \bigl(M,(i^\star,j^\star,D_{i^\star,j^\star})\bigr)$
    \State $I \gets I \setminus \{i^\star\}$
    \State $J \gets J \setminus \{j^\star\}$
\EndWhile

\State $\bar d \gets 
       \frac{1}{|M|}
       \displaystyle\sum_{(i,j,d)\in M} d$
\State \Return $(M,\bar d)$

\end{algorithmic}
\end{algorithm}

When the number of source shapelets is smaller than the number of target shapelets, Algorithm~\ref{alg:sm_shapelet} repeats the source shapelet list before greedy matching.
The duplicated entries are treated as distinct matching candidates, which allows every target-side discriminative shapelet to contribute to the final distance.

\subsection{Comparison between Dataset Similarity Metrics and Transferability Estimation Methods}

\begin{table*}[h!]
    \caption{Comparison of source-selection methods in terms of training requirement and average computational time (seconds).
    The last three columns report the incremental cost of adding a new backbone architecture, one candidate source dataset, or one target dataset under our experimental setting.
    Actual computation times may vary across datasets.}
    \label{tab:comparison}
    \centering
    \makebox[\linewidth][c]{%
    \begin{tabular}{lccccc}
    \toprule
    \textbf{Method} & \textbf{Type} & \textbf{Training} & \textbf{Per New Model} & \textbf{\shortstack[c]{Per New \\ Source Dataset}}  & \textbf{\shortstack[c]{Per New \\ Target Dataset}}\\
    \midrule
    \multicolumn{5}{l}{\textit{Transferability Estimation}} &  \\
    H-score   & Feature-based    & \cmark & 36,000 s (GPU) & 280 s (GPU) & 170 s (GPU) \\
    LEEP      & Prediction-based & \cmark & 36,000 s (GPU) & 280 s (GPU) & 170 s (GPU) \\
    LogME     & Feature-based    & \cmark & 36,000 s (GPU) & 280 s (GPU)  & 170 s (GPU)\\
    NCE       & Prediction-based & \cmark & 36,000 s (GPU) & 280 s (GPU) & 170 s (GPU) \\
    TransRate & Feature-based    & \cmark & 36,000 s (GPU) & 280 s (GPU) & 170 s (GPU) \\
    TMI       & Feature-based    & \cmark & 36,000 s (GPU) & 280 s (GPU) & 170 s (GPU) \\
    \midrule
    \multicolumn{5}{l}{\textit{Dataset Similarity Metrics}} &  \\
    DBA-DTW              & Global similarity & \xmark & --- & 52 s (CPU only) & 25,000 s (CPU only) \\
    Avg. Shapelet (Ours) & Local similarity  & \xmark & --- & 98 s (CPU only) & 110 s (CPU only) \\
    Min. Shapelet (Ours) & Local similarity  & \xmark & --- & 98 s (CPU only) & 110 s (CPU only) \\
    Proposed SM (Ours)   & Local similarity  & \xmark & --- & 98 s (CPU only) & 110 s (CPU only) \\
    \bottomrule
    \end{tabular}}
\end{table*}

As summarized in Table~\ref{tab:comparison}, source-selection methods can be divided into transferability estimation methods and dataset similarity metrics.
Transferability estimation methods are model dependent: each new backbone or source dataset requires source-model pre-training, and each new target must be evaluated through the trained source models. 
In our setting, this required about 36,000~s for one new backbone, 280~s for one new source dataset, and 170~s for one new target dataset.

Dataset similarity metrics avoid source-model pre-training because they compare source and target datasets directly. 
DBA-DTW is also training-free, but its target-side comparison is costly because DBA prototypes must be compared with DTW, requiring about 25,000~s per new target in our setting. 
The proposed shapelet-based metrics require one-time shapelet extraction, about 98~s per new source dataset and 110~s per new target dataset, after which source--target scores are computed using Euclidean distances between discovered shapelets. 
Thus, the ranking can be computed once without training a backbone and reused as an input to different architectures, although the downstream benefit of that ranking remains architecture-dependent.

\section{Experimental Results}
\subsection{Dataset}
We evaluated the proposed method using 128 univariate time series datasets from the UCR Archive~\cite{UCRArchive2018}\footnote{\url{https://www.cs.ucr.edu/\%7Eeamonn/time_series_data_2018/}}.
We use the division of training and test sets as determined by the archive.
Except for the temporal-length resizing described in Section~\ref{sec:multi}, we used the archive values as released and did not apply additional z-normalization. 
Importantly, the UCR 2018 archive is not uniformly normalized: the legacy Summer 2015 datasets are z-normalized, whereas the Fall 2018 additions are kept in their original scale unless they were already normalized by the data donor. Therefore, the normalization status depends on the dataset.
Table~\ref{tab:ucr} shows characteristics of datasets in UCR Archive.

\begin{table}[t]
\centering
\caption{\label{tab:ucr}Summary of the 128 univariate UCR datasets used in the experiments.}
\begin{tabular}{lccc}
\toprule
Property & Min & Median & Max \\
\midrule
Series length & 15 & 344 & 2,844 \\
Number of classes & 2 & 4 & 60 \\
Training samples & 16 & 190.5 & 8,926 \\
Test samples & 20 & 316 & 16,800 \\
\bottomrule
\end{tabular}
\end{table}

\subsection{Settings and Architecture}
We evaluated our proposed method on three representative architectures for time series classification: CNNs, ViT and PatchTST. For CNNs, we adopted the VGG~\cite{simonyan2015deep}-based architecture with three blocks of convolutional layers and pooling layers.
Max pooling follows the first two blocks, and global average pooling (GAP) follows the final block. 
We used GAP to obtain a fixed-dimensional representation across
datasets of different temporal lengths and to provide a common
interface for the evaluated transferability estimators.

We implemented two patch-based Transformer models: ViT, originally introduced for image recognition~\cite{dosovitskiy2020image}, and PatchTST, adapted to time-series tasks~\cite{nie2022time}.
Each model's architecture is configured with three Transformer encoder blocks, a model dimension of 16, four attention heads, and a feed-forward network dimension of 128.
The input data is segmented into patches of length 16; ViT uses non-overlapping patches (stride 16), whereas PatchTST uses 50\%-overlapping patches (stride 8).
For the final classification task, a head consisting of a flatten layer followed by a linear classification layer with a softmax activation function is appended to the encoder output.

\begin{table}[t]
\centering
\caption{Training and source-selection hyperparameters used in the
experiments. The shapelet length and number of candidate shapelets
follow the preliminary study, whereas $K=16$ is specific to the
present experiments. Fine-tuning was repeated three times from a
fixed pre-trained checkpoint. A dash indicates that the setting is
not applicable.}
\label{tab:hyperparameters}

\begingroup
\small
\setlength{\tabcolsep}{5pt}
\renewcommand{\arraystretch}{1.05}

\begin{tabular}{lccc}
\toprule
\textbf{Setting} &
\textbf{VGG} &
\textbf{ViT} &
\textbf{PatchTST} \\
\midrule
Pre-training iterations
    & 10,000 & 20,000 & 20,000 \\
Fine-tuning iterations
    & 5,000 & 10,000 & 10,000 \\
Optimizer
    & Adam & Adam & Adam \\
Learning rate
    & $10^{-4}$ & $10^{-4}$ & $10^{-4}$ \\
Batch size
    & 32 & 32 & 32 \\
Fine-tuning repetitions
    & 3 & 3 & 3 \\
Patch size
    & -- & 16 & 16 \\
Shapelet length $\ell_s$
    & 15 & 15 & 15 \\
Candidate shapelets per class $Q$
    & 10 & 10 & 10 \\
Selected sources $K$ (multi-source)
    & 16 & 16 & 16 \\
\bottomrule
\end{tabular}

\endgroup
\end{table}

Training and source-selection settings are summarized in
Table~\ref{tab:hyperparameters}. Following our preliminary
study~\cite{lee2024model}, we set the shapelet length to 15 and
retained 10 candidate shapelets per class. The preliminary study
used $K=14$, whereas the present study selected $K=16$ as a
representative operating point within the near-saturation region
observed in the exploratory VGG analysis in
Fig.~\ref{fig:nb_datasets}. The same value was used across all
source-selection methods and backbones. We do not claim that
$K=16$ is uniquely optimal, and comparisons at this setting should
be interpreted in light of this exploratory choice.

\subsection{Comparative Evaluation}
We compare three learning strategies: no transfer learning, single-source transfer learning, and multi-source transfer learning with $K=16$ selected sources. The source selection criteria include conventional transferability estimators (H-score~\cite{Bao_2019}, LEEP~\cite{pmlr-v119-nguyen20b}, LogME~\cite{pmlr-v139-you21b}, NCE~\cite{Tran_2019}, TransRate~\cite{pmlr-v162-huang22d}, and TMI~\cite{xu2023fast}), the DBA-DTW dataset similarity baseline~\cite{Ismail_Fawaz_2018}, and the proposed shapelet-based measures.
Results are reported as mean $\pm$ standard deviation over three fine-tuning runs from a fixed pre-trained checkpoint. 
For both Tables~\ref{tab:results_cnn} and~\ref{tab:results_transformer_patchtst}, the 95\% percentile bootstrap confidence intervals were computed using 10,000 resamples of the 128 target datasets. 
The three fine-tuning runs were averaged within each target dataset before resampling, and the same bootstrap resamples were used across all methods. For significance testing, Holm correction was applied in both Tables~\ref{tab:results_cnn} and~\ref{tab:results_transformer_patchtst}.
The reported standard deviation is the average of the per-target standard deviations computed over these three runs; it is not the standard deviation of the mean accuracies across the 128 target datasets.
The paired statistical unit is the target dataset, and corrected paired comparisons for the main $K=16$ setting are reported in Table~\ref{tab:main-effect-size-k16}.
Paired $t$-tests were conducted on the 128 dataset-level paired differences after averaging the three runs within each dataset.
We used the paired $t$-test as the primary analysis because the statistical quantity of interest is the mean paired difference and the number of paired datasets is relatively large ($n=128$).
To assess the robustness of the conclusions to the parametric assumptions of the $t$-test, we additionally performed dataset-level bootstrap confidence interval estimation and Wilcoxon signed-rank sensitivity analyses.
These analyses yielded the same qualitative conclusions for the primary comparisons between Proposed SM and no transfer learning.

   \begin{table}[h!]
      \caption{Average classification accuracy (\%) across the 128 UCR target datasets for the VGG backbone. Single-source transfer uses the top-ranked source dataset, whereas multi-source transfer uses the top-$K$ sources with $K=16$.
      Each entry reports the across-target mean accuracy ($\pm$) the average within-target standard deviation over three fine-tuning runs, followed by a 95\% percentile bootstrap confidence interval in brackets. Methods are grouped according to whether they require source-model training or use training-free dataset similarity. Boldface indicates the best result in each column, and asterisks indicate significance relative to no transfer learning based on Holm-adjusted paired $t$-tests.}
      \label{tab:results_cnn}
      \centering
      \small
      \renewcommand{\arraystretch}{1.25}
      \begin{tabular}{lcc}
      \toprule
      \textbf{Method} & \textbf{Single} & \textbf{Multi} \\
      \midrule
      \multicolumn{3}{l}{\textit{Baseline: No Transfer Learning}} \\
      & \multicolumn{2}{c}{
          \shortstack{
              74.18 $\pm$ 3.60\\[-1pt]
              {\scriptsize [70.78, 77.51]}
          }
        } \\
      \midrule
      \multicolumn{3}{l}{\textit{Transferability Metrics}} \\

      H-score
      & \shortstack{
          77.92 $\pm$ 2.11\textsuperscript{***}\\[-1pt]
          {\scriptsize [74.69, 80.96]}
        }
      & \shortstack{
          80.05 $\pm$ 1.43\textsuperscript{***}\\[-1pt]
          {\scriptsize [77.05, 82.88]}
        } \\

      LEEP
      & \shortstack{
          77.73 $\pm$ 2.05\textsuperscript{***}\\[-1pt]
          {\scriptsize [74.47, 80.85]}
        }
      & \shortstack{
          80.12 $\pm$ 1.60\textsuperscript{***}\\[-1pt]
          {\scriptsize [77.14, 82.95]}
        } \\

      LogME
      & \shortstack{
          79.17 $\pm$ 1.89\textsuperscript{***}\\[-1pt]
          {\scriptsize [75.93, 82.15]}
        }
      & \shortstack{
          79.98 $\pm$ 1.78\textsuperscript{***}\\[-1pt]
          {\scriptsize [76.93, 82.80]}
        } \\

      NCE
      & \shortstack{
          78.75 $\pm$ 1.68\textsuperscript{***}\\[-1pt]
          {\scriptsize [75.42, 81.82]}
        }
      & \shortstack{
          80.22 $\pm$ 1.44\textsuperscript{***}\\[-1pt]
          {\scriptsize [77.23, 83.04]}
        } \\

      TransRate
      & \shortstack{
          77.03 $\pm$ 1.90\textsuperscript{**}\\[-1pt]
          {\scriptsize [73.67, 80.18]}
        }
      & \shortstack{
          79.68 $\pm$ 1.61\textsuperscript{***}\\[-1pt]
          {\scriptsize [76.67, 82.49]}
        } \\

      TMI
      & \shortstack{
          72.21 $\pm$ 2.32\\[-1pt]
          {\scriptsize [67.99, 76.18]}
        }
      & \shortstack{
          80.18 $\pm$ 1.56\textsuperscript{***}\\[-1pt]
          {\scriptsize [77.25, 82.90]}
        } \\

      \midrule
      \multicolumn{3}{l}{\textit{Global Dataset Similarity Metrics}} \\

      DBA-DTW
      & \shortstack{
          78.42 $\pm$ 2.12\textsuperscript{***}\\[-1pt]
          {\scriptsize [75.17, 81.50]}
        }
      & \shortstack{
          80.41 $\pm$ 1.67\textsuperscript{***}\\[-1pt]
          {\scriptsize [77.37, 83.22]}
        } \\

      \midrule
      \multicolumn{3}{l}{\textit{Local Shapelet Similarity Metrics}} \\

      Avg. Shapelet (Ours)
      & \shortstack{
          76.47 $\pm$ 2.12\textsuperscript{**}\\[-1pt]
          {\scriptsize [73.15, 79.60]}
        }
      & \shortstack{
          79.94 $\pm$ 1.64\textsuperscript{***}\\[-1pt]
          {\scriptsize [77.01, 82.70]}
        } \\

      Min. Shapelet (Ours)
      & \shortstack{
          78.70 $\pm$ 1.65\textsuperscript{***}\\[-1pt]
          {\scriptsize [75.50, 81.70]}
        }
      & \shortstack{
          79.97 $\pm$ 1.79\textsuperscript{***}\\[-1pt]
          {\scriptsize [76.90, 82.80]}
        } \\

      Proposed SM (Ours)
      & \shortstack{
          78.32 $\pm$ 1.96\textsuperscript{***}\\[-1pt]
          {\scriptsize [75.18, 81.32]}
        }
      & \shortstack{
          \textbf{80.78} $\pm$ 1.69\textsuperscript{***}\\[-1pt]
          {\scriptsize [77.87, 83.49]}
        } \\

      \bottomrule
      \multicolumn{3}{r}{
          \scriptsize
          * $p_{\mathrm{Holm}}<0.05$,
          ** $p_{\mathrm{Holm}}<0.01$,
          *** $p_{\mathrm{Holm}}<0.001$
      }\\
      \end{tabular}
  \end{table}

  \begin{table}[h!]
      \caption{Average classification accuracy (\%) across the 128 UCR target datasets for the ViT and PatchTST backbones. Single-source transfer uses the top-ranked source dataset, whereas multi-source transfer uses the top-$K$ sources with $K=16$. Each entry reports the across-target mean accuracy ($\pm$) the average within-target standard deviation over three fine-tuning runs, followed by a 95\% percentile bootstrap confidence interval in brackets. Methods are grouped according to whether they require source-model training or use training-free dataset similarity. Boldface indicates the best result within each backbone and transfer setting, and asterisks indicate significance relative to no transfer learning based on Holm-adjusted paired $t$-tests.}
      \label{tab:results_transformer_patchtst}
      \centering
      \small
      \renewcommand{\arraystretch}{1.25}
      \makebox[\linewidth][c]{%
      \begin{tabular}{lcccc}
      \toprule
      & \multicolumn{2}{c}{\textbf{ViT}}
      & \multicolumn{2}{c}{\textbf{PatchTST}} \\
      \cmidrule(lr){2-3} \cmidrule(lr){4-5}
      \textbf{Method}
      & \textbf{Single}
      & \textbf{Multi}
      & \textbf{Single}
      & \textbf{Multi} \\
      \midrule
      \multicolumn{5}{l}{\textit{Baseline: No Transfer Learning}} \\
      & \multicolumn{2}{c}{
          \shortstack{
              70.22 $\pm$ 2.01\\[-1pt]
              {\scriptsize [66.44, 73.85]}
          }
        }
      & \multicolumn{2}{c}{
          \shortstack{
              73.86 $\pm$ 2.34\\[-1pt]
              {\scriptsize [70.43, 77.12]}
          }
        } \\
      \midrule
      \multicolumn{5}{l}{\textit{Transferability Metrics}} \\

      H-score
      & \shortstack{
          67.06 $\pm$ 3.46\textsuperscript{**}\\[-1pt]
          {\scriptsize [63.02, 70.93]}
        }
      & \shortstack{
          71.07 $\pm$ 1.41\\[-1pt]
          {\scriptsize [67.36, 74.55]}
        }
      & \shortstack{
          66.39 $\pm$ 4.24\textsuperscript{***}\\[-1pt]
          {\scriptsize [62.56, 70.06]}
        }
      & \shortstack{
          75.20 $\pm$ 1.71\textsuperscript{*}\\[-1pt]
          {\scriptsize [71.92, 78.36]}
        } \\

      LEEP
      & \shortstack{
          71.69 $\pm$ 1.46\textsuperscript{**}\\[-1pt]
          {\scriptsize [67.95, 75.25]}
        }
      & \shortstack{
          71.09 $\pm$ 2.00\\[-1pt]
          {\scriptsize [67.40, 74.58]}
        }
      & \shortstack{
          75.65 $\pm$ 1.92\\[-1pt]
          {\scriptsize [72.17, 78.97]}
        }
      & \shortstack{
          76.86 $\pm$ 1.72\textsuperscript{***}\\[-1pt]
          {\scriptsize [73.58, 80.00]}
        } \\

      LogME
      & \shortstack{
          68.73 $\pm$ 1.64\textsuperscript{*}\\[-1pt]
          {\scriptsize [65.02, 72.24]}
        }
      & \shortstack{
          70.29 $\pm$ 1.55\\[-1pt]
          {\scriptsize [66.47, 73.90]}
        }
      & \shortstack{
          74.06 $\pm$ 1.82\\[-1pt]
          {\scriptsize [70.45, 77.42]}
        }
      & \shortstack{
          75.52 $\pm$ 1.99\textsuperscript{**}\\[-1pt]
          {\scriptsize [72.20, 78.69]}
        } \\

      NCE
      & \shortstack{
          71.50 $\pm$ 1.52\textsuperscript{*}\\[-1pt]
          {\scriptsize [67.83, 75.00]}
        }
      & \shortstack{
          71.26 $\pm$ 1.73\\[-1pt]
          {\scriptsize [67.45, 74.84]}
        }
      & \shortstack{
          75.81 $\pm$ 1.77\textsuperscript{*}\\[-1pt]
          {\scriptsize [72.41, 79.06]}
        }
      & \shortstack{
          76.62 $\pm$ 1.93\textsuperscript{***}\\[-1pt]
          {\scriptsize [73.38, 79.71]}
        } \\

      TransRate
      & \shortstack{
          70.33 $\pm$ 1.65\\[-1pt]
          {\scriptsize [66.59, 73.90]}
        }
      & \shortstack{
          71.75 $\pm$ 1.46\textsuperscript{***}\\[-1pt]
          {\scriptsize [68.09, 75.22]}
        }
      & \shortstack{
          75.18 $\pm$ 1.84\\[-1pt]
          {\scriptsize [71.82, 78.40]}
        }
      & \shortstack{
          76.91 $\pm$ 1.69\textsuperscript{***}\\[-1pt]
          {\scriptsize [73.68, 79.99]}
        } \\

      TMI
      & \shortstack{
          68.30 $\pm$ 1.86\textsuperscript{*}\\[-1pt]
          {\scriptsize [64.50, 71.88]}
        }
      & \shortstack{
          70.78 $\pm$ 1.51\\[-1pt]
          {\scriptsize [67.06, 74.32]}
        }
      & \shortstack{
          72.96 $\pm$ 2.10\\[-1pt]
          {\scriptsize [69.63, 76.17]}
        }
      & \shortstack{
          75.85 $\pm$ 1.69\textsuperscript{*}\\[-1pt]
          {\scriptsize [72.55, 78.97]}
        } \\

      \midrule
      \multicolumn{5}{l}{\textit{Global Dataset Similarity Metrics}} \\

      DBA-DTW
      & \shortstack{
          71.94 $\pm$ 1.59\textsuperscript{**}\\[-1pt]
          {\scriptsize [68.31, 75.42]}
        }
      & \shortstack{
          \textbf{72.38} $\pm$ 1.72\textsuperscript{***}\\[-1pt]
          {\scriptsize [68.76, 75.77]}
        }
      & \shortstack{
          75.83 $\pm$ 1.86\textsuperscript{*}\\[-1pt]
          {\scriptsize [72.48, 79.09]}
        }
      & \shortstack{
          \textbf{77.30} $\pm$ 1.59\textsuperscript{***}\\[-1pt]
          {\scriptsize [74.04, 80.36]}
        } \\

      \midrule
      \multicolumn{5}{l}{\textit{Local Shapelet Similarity Metrics}} \\

      Avg. Shapelet (Ours)
      & \shortstack{
          69.81 $\pm$ 1.52\\[-1pt]
          {\scriptsize [66.10, 73.38]}
        }
      & \shortstack{
          71.22 $\pm$ 1.48\textsuperscript{*}\\[-1pt]
          {\scriptsize [67.52, 74.74]}
        }
      & \shortstack{
          73.26 $\pm$ 2.15\\[-1pt]
          {\scriptsize [69.87, 76.54]}
        }
      & \shortstack{
          76.48 $\pm$ 1.68\textsuperscript{**}\\[-1pt]
          {\scriptsize [73.19, 79.67]}
        } \\

      Min. Shapelet (Ours)
      & \shortstack{
          70.65 $\pm$ 1.39\\[-1pt]
          {\scriptsize [67.00, 74.15]}
        }
      & \shortstack{
          71.09 $\pm$ 1.53\\[-1pt]
          {\scriptsize [67.41, 74.56]}
        }
      & \shortstack{
          75.75 $\pm$ 1.86\\[-1pt]
          {\scriptsize [72.48, 78.92]}
        }
      & \shortstack{
          76.22 $\pm$ 1.78\textsuperscript{**}\\[-1pt]
          {\scriptsize [72.97, 79.34]}
        } \\

      Proposed SM (Ours)
      & \shortstack{
          71.16 $\pm$ 1.50\textsuperscript{*}\\[-1pt]
          {\scriptsize [67.50, 74.59]}
        }
      & \shortstack{
          71.88 $\pm$ 1.69\textsuperscript{***}\\[-1pt]
          {\scriptsize [68.22, 75.36]}
        }
      & \shortstack{
          75.43 $\pm$ 1.62\\[-1pt]
          {\scriptsize [72.08, 78.57]}
        }
      & \shortstack{
          76.65 $\pm$ 1.92\textsuperscript{***}\\[-1pt]
          {\scriptsize [73.37, 79.80]}
        } \\

      \bottomrule
      \multicolumn{5}{r}{
          \scriptsize
          * $p_{\mathrm{Holm}}<0.05$,
          ** $p_{\mathrm{Holm}}<0.01$,
          *** $p_{\mathrm{Holm}}<0.001$
      }\\
      \end{tabular}}
  \end{table}

Tables~\ref{tab:results_cnn} and~\ref{tab:results_transformer_patchtst} show two main trends.
First, multi-source transfer learning generally improves over the corresponding single-source setting, indicating that aggregating several selected sources reduces dependence on a single potentially mismatched source. On VGG, the proposed Shapelet Matching (SM) obtains the best average multi-source accuracy, improving from the no-transfer baseline of 74.18\% to 80.78\%.
Second, the best source-selection criterion depends on the backbone architecture. For ViT and PatchTST, DBA-DTW achieves the highest mean multi-source accuracy, whereas the proposed SM remains competitive.
This result suggests that shapelet-based source selection is especially well aligned with the VGG-style CNN, while patch-based Transformer models may benefit more from global similarity criteria.

For Proposed SM, single-source transfer yielded lower accuracy than no transfer learning on 36, 52, and 60 of the 128 target datasets for VGG, ViT, and PatchTST, respectively.
Under multi-source transfer, these numbers decreased to 21, 40, and 45, corresponding to net reductions of 15, 12, and 15 targets.
Multi-source SM also outperformed the corresponding single-source setting on 82 target datasets for each backbone, although the mean improvement was not statistically significant for ViT.
Thus, multi-source transfer improved average accuracy while reducing the risk of negative transfer.

\begin{table}[h!]
\centering
\caption{Paired accuracy comparisons across 128 UCR target datasets.
For each backbone, differences are computed per target dataset as Proposed ($K=16$, multi-source transfer with Shapelet Matching) minus the comparator and then summarized across datasets.
CIs are percentile-bootstrap 95\% intervals, and $p$-values are from two-sided paired $t$-tests with Holm adjustment applied jointly across all 12 comparisons shown (four comparisons for each of the three backbones).
W/L/T denotes wins, losses, and ties.}

\label{tab:main-effect-size-k16}
\makebox[\linewidth][c]{%
\begin{tabular}{lcccccc}
\toprule
Comparison & Mean $\Delta$ acc. (pp) & 95\% CI & Holm $p$ &
Cohen's $d_z$ & Rank-biserial & W/L/T \\
\midrule
\multicolumn{7}{l}{\textit{VGG}} \\
Proposed - No TL & 6.60 & [5.03, 8.26] & $<$0.001 & 0.71 & 0.80 & 103/21/4 \\
Proposed - DBA-DTW & 0.37 & [-0.11, 0.87] & 0.462 & 0.13 & 0.15 & 62/54/12 \\
Proposed - Minimum Shapelet & 0.81 & [0.31, 1.36] & 0.021 & 0.27 & 0.25 & 71/48/9 \\
Proposed - Single-source (SM) & 2.47 & [1.44, 3.64] & $<$0.001 & 0.39 & 0.53 & 82/38/8 \\
\midrule
\multicolumn{7}{l}{\textit{ViT}} \\
Proposed - No TL & 1.66 & [0.84, 2.50] & 0.001 & 0.35 & 0.46 & 81/40/7 \\
Proposed - DBA-DTW & -0.50 & [-1.28, 0.18] & 0.462 & -0.12 & -0.05 & 63/59/6 \\
Proposed - Minimum Shapelet & 0.79 & [0.22, 1.34] & 0.042 & 0.24 & 0.40 & 83/36/9 \\
Proposed - Single-source (SM) & 0.72 & [-0.14, 1.53] & 0.462 & 0.15 & 0.36 & 82/40/6 \\
\midrule
\multicolumn{7}{l}{\textit{PatchTST}} \\
Proposed - No TL & 2.79 & [1.52, 4.26] & 0.001 & 0.35 & 0.43 & 77/45/6 \\
Proposed - DBA-DTW & -0.65 & [-1.42, 0.13] & 0.462 & -0.14 & -0.15 & 60/62/6 \\
Proposed - Minimum Shapelet & 0.43 & [-0.14, 1.02] & 0.462 & 0.13 & 0.16 & 69/53/6 \\
Proposed - Single-source (SM) & 1.22 & [0.47, 1.97] & 0.015 & 0.28 & 0.42 & 82/41/5 \\
\bottomrule
  \multicolumn{7}{l}{\scriptsize
  \textit{Note.} As descriptive benchmarks, absolute Cohen's $d_z$ values
  of approximately 0.2, 0.5, and 0.8 are conventionally considered small,
  medium, and large, respectively.}\\
\end{tabular}}
\end{table}

Table~\ref{tab:main-effect-size-k16} provides the corrected paired analysis for the main claims.
The proposed multi-source SM significantly improves over no transfer learning for all three backbones after Holm correction.
The effect is largest for VGG ($\Delta=6.60$ percentage points, 95\% CI [5.03, 8.26], $p<0.001$, Cohen's $d_z=0.71$), and smaller but still significant for ViT and PatchTST.
The W/L/T counts further show that the VGG improvement is broadly distributed across target datasets.

The comparison with alternative selection strategies gives a more critical view.
On VGG, the proposed SM significantly improves over Minimum Shapelet, supporting the benefit of target-guided dense shapelet matching.
However, the difference from DBA-DTW is not statistically significant for any backbone, and DBA-DTW obtains slightly higher mean accuracy on the two Transformer-based models.
Therefore, the proposed method should not be interpreted as universally superior in accuracy.
Its main advantage is a training-free, inspectable source-selection criterion whose rankings can be reused across backbones.

\section{Discussion}
\subsection{Effect of Number of Source Datasets in Multi-Source Transfer Learning}
\begin{figure}[ht]
    \centering
    \includegraphics[width=0.8\linewidth]{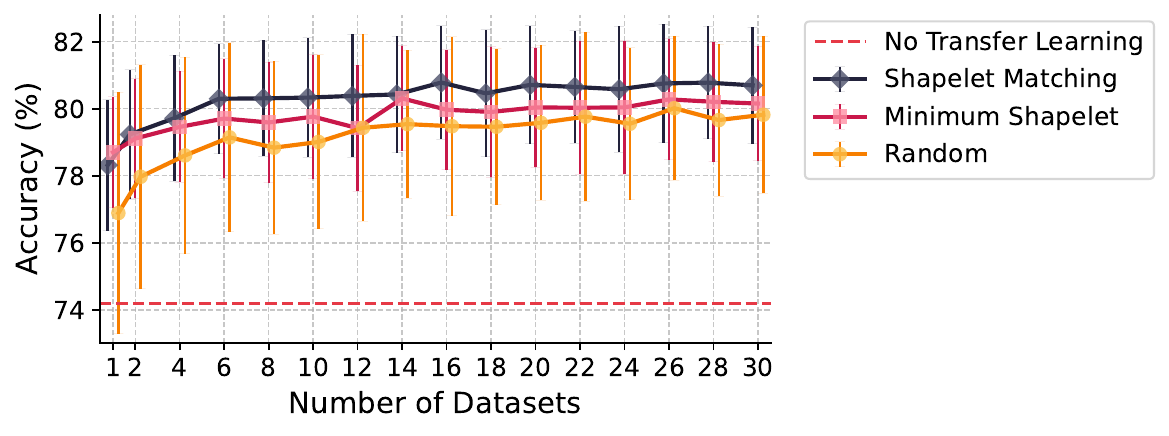}
    \caption{Effect of the number $K$ of selected source datasets on VGG multi-source transfer learning. The y-axis reports the mean classification accuracy over the 128 UCR target datasets, and the dashed horizontal line indicates the no-transfer baseline. Error bars indicate the average per-target standard deviation over three fine-tuning runs.}
    \label{fig:nb_datasets}
\end{figure}

Fig.~\ref{fig:nb_datasets} shows that performance enters a near-saturation region at approximately $K=14$.
Although the exact accuracy fluctuates across neighboring values, increasing $K$ beyond this region provides only marginal additional benefit.
Under the fixed training budget, increasing $K$ beyond this region yields limited additional accuracy gains in the evaluated setting.

\subsection{Connecting Minimum and Shapelet Matching Approaches}

\begin{figure}[ht]
    \centering
    \includegraphics[width=0.65\linewidth]{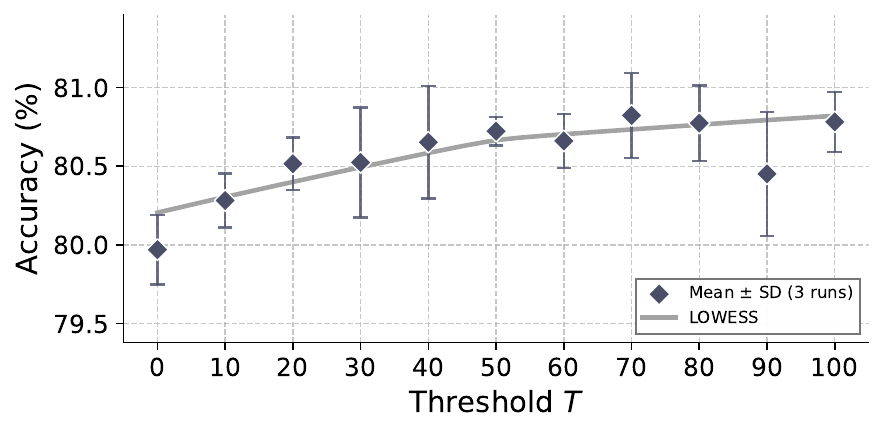}
    \caption{Performance transition from Minimum Shapelet to Shapelet Matching on the VGG backbone under the $K=16$ multi-source setting. 
    The point at threshold 0 denotes the Minimum Shapelet, which uses the single closest shapelet pair.
    For each positive threshold $T$, the score averages the smallest $\lfloor (T/100)M \rfloor$ distances among the $M$ one-to-one pairs produced by the greedy matching procedure; at threshold 100, all matched pairs are included, yielding full Shapelet Matching as described in Algorithm~\ref{alg:sm_shapelet}.
    For each run, classification accuracy is averaged over the 128 UCR target datasets; diamonds and error bars show the mean and standard deviation across three runs, respectively.
    The gray dotted curve represents LOWESS smoothing ($\mathrm{frac}=1.0$, three robust reweighting iterations) and is included only as a visual guide.} 
    \label{fig:threshold}
\end{figure}

Fig.~\ref{fig:threshold} presents a sensitivity analysis of the matching-coverage threshold, which controls the extent of source--target shapelet matching.
Threshold 0 retains only the single closest pair and thus corresponds to Minimum Shapelet, whereas threshold 100 uses the full set of pairs obtained through the greedy matching.
The mean accuracies at intermediate thresholds are non-monotonic.
The endpoint accuracy increases from 79.97\% for Minimum Shapelet to 80.78\% for full Shapelet Matching.
The highest observed mean is 80.82\% at threshold 70, only 0.04 percentage points above full matching.
We use full matching in the main experiments without introducing an additional coverage-tuning parameter.
The LOWESS curve is included only as a visual guide.

\subsection{Architecture-dependent Performance}
\label{sec:MD}

\begin{figure}[ht]
    \centering
    \includegraphics[width=\linewidth]{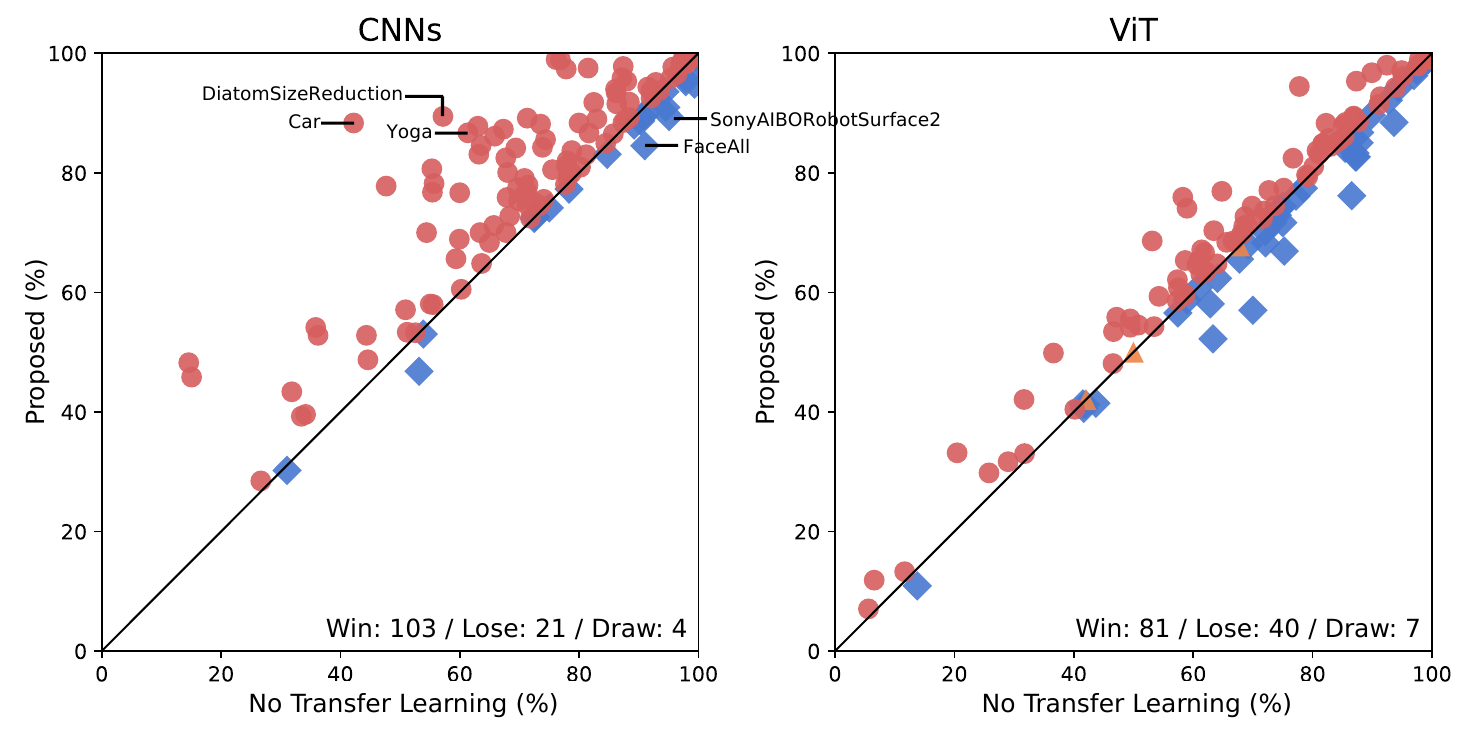}
    \caption{Dataset-wise comparison between the proposed Shapelet Matching multi-source transfer learning and no transfer learning. Each point represents one UCR target dataset and reports the average accuracy over three fine-tuning runs. The x-axis shows the no-transfer accuracy, and the y-axis shows the accuracy obtained by the proposed method; points above the diagonal indicate improvement and points below the diagonal indicate degradation. Red points indicate wins for the proposed method, and the win/loss/draw counts summarize the 128 target datasets for each displayed backbone.}
    
    \label{fig:winlose}
\end{figure}

Fig.~\ref{fig:winlose} shows the dataset-wise comparisons with no transfer learning for the VGG and ViT backbones.
The benefit of the proposed multi-source transfer pipeline varies across the evaluated backbones.
Relative to no transfer learning, Proposed SM yields mean accuracy gains of 6.60, 1.66, and 2.79 percentage points for VGG, ViT, and PatchTST, respectively, with win/loss/tie counts of 103/21/4, 81/40/7, and 77/45/6 (Table~\ref{tab:main-effect-size-k16}).
The largest observed gain is therefore obtained with the VGG-based CNN.
However, Proposed SM does not significantly outperform DBA-DTW on any of the three backbones.

One plausible explanation is that shapelet-based ranking emphasizes local discriminative subsequences that may be particularly compatible with the VGG backbone.
However, the present experiments do not isolate this mechanism or establish that patch tokenization causes the smaller gains observed for the Transformers.
These findings describe architecture-dependent behavior under the evaluated configurations and should not be generalized to all CNN or Transformer architectures.

\subsection{Dataset-Level Trends and Failure Cases}
  \begin{figure}[ht]
      \centering
      \includegraphics[width=0.55\linewidth]{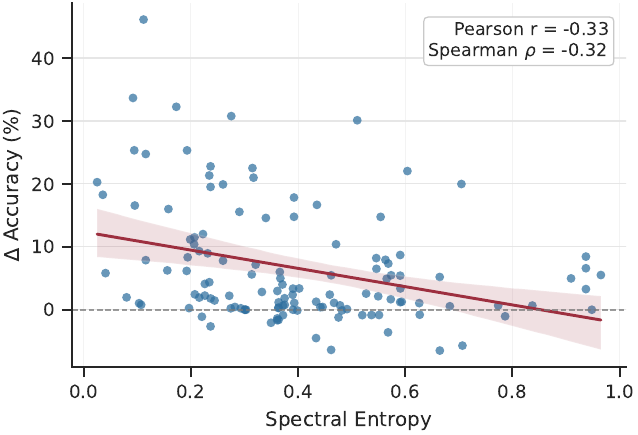}
      \caption{
      Accuracy gain of Shapelet Matching over no transfer learning versus normalized spectral entropy across 128 UCR datasets (VGG, $K=16$). Each point represents a target dataset; the red line and shading show the linear regression fit and its 95\% confidence interval. Lower (higher) entropy indicates spectral power concentrated in fewer (spread across more) frequencies.}
      \label{fig:spectral_entropy_delta}
  \end{figure}

  \begin{figure}[h!]
      \centering
      \includegraphics[width=\linewidth]{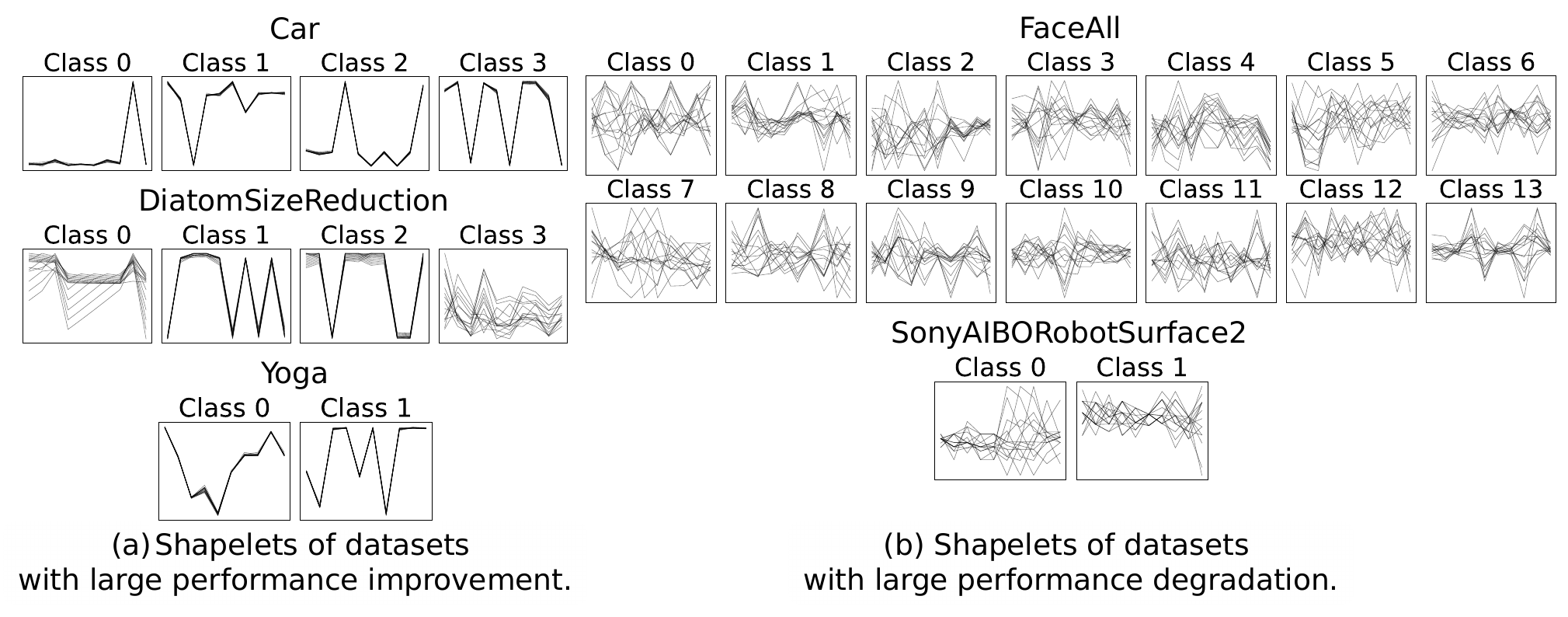}
      \caption{Illustrative examples of discriminative shapelets extracted from datasets with large improvement or degradation under the proposed method.
      Each small plot shows the discovered shapelets for one class of the corresponding UCR dataset.
      Panel (a) shows datasets where Shapelet Matching improved over no transfer learning, whereas panel (b) shows datasets where it degraded.
      These examples suggest that the proposed method is more effective when compact local class-discriminative subsequences are visible, and less effective when such local patterns are weak or ambiguous.}
      \label{fig:shapelet_ex}
  \end{figure}

The effectiveness of the proposed shapelet-based transfer learning method depends on the dataset--method fit, especially whether discriminative and class-representative local subsequences can be extracted.
To examine this dependence, we analyze normalized PSD-based spectral entropy and qualitative examples of the discovered shapelets.

For each time series in the training split of each target dataset, we estimated the PSD using Welch’s method, excluded the DC bin, and normalized the PSD values at the remaining frequency bins to sum to one. 
Let $B$ denote the number of retained frequency bins and $p_b$ the normalized PSD value at bin $b$.
We then computed the normalized spectral entropy as $H_{\mathrm{norm}}=-\frac{\sum_{b=1}^{B}p_b\log p_b}{\log B},$ and averaged $H_{\mathrm{norm}}$ across the training series for each target dataset.

Fig.~\ref{fig:spectral_entropy_delta} shows the relationship between spectral entropy and the dataset-wise accuracy gain of Shapelet Matching over no transfer learning under the VGG multi-source setting with $K=16$.
The association is moderately negative, with Pearson $r=-0.33$ and Spearman $\rho=-0.32$, suggesting that larger gains tend to occur when target datasets have more concentrated and regular spectral structure.
However, the presence of outliers indicates that spectral entropy is only a partial predictor of transfer performance.

The qualitative examples in Fig.~\ref{fig:shapelet_ex} support this interpretation.
Datasets with large positive gains often exhibit compact and visually consistent shapelets, whereas datasets with degradation tend to show weak, ambiguous, or poorly localized patterns.
These cases suggest that the target decision boundary may depend more on global shape, long-range temporal structure, or heterogeneous signal characteristics than on compact local subsequences.

Shapelet Matching tends to be more beneficial when the target exhibits clear spectral structure and extractable local subsequences, but spectral entropy alone is insufficient to predict transfer performance.
When the data are spectrally complex or the discriminative information is not well represented by local shapelets, the method may select sources that appear locally similar but do not provide useful initialization for the target task.

\subsection{Relation to Recent Time Series Foundation Models}
Recent time-series foundation models shift the focus from dataset-specific transfer to large-scale pre-training and adaptation.
Our method addresses a different problem: given a target classification dataset and a finite pool of heterogeneous labeled sources, it identifies sources that contain target-relevant local discriminative subsequences.
Its advantages are that it requires no separate source-model training, produces source rankings reusable across backbones, and remains potentially interpretable through matched shapelets.
These properties are useful when large-scale foundation-model pre-training or released checkpoints are unavailable.
However, Shapelet Matching does not learn a universal representation, does not provide zero-shot inference, and depends on the existence of transferable local shapelets.
It may therefore be less suitable when the target task is governed mainly by long-range temporal structure, global shape, or multivariate interactions.
Future work should investigate how shapelet-based source ranking can support foundation-model adaptation, for example by selecting data for continued pre-training or by providing interpretable transfer diagnostics.

\section{Limitations}
\label{sec:lim}
The focus of this study was confined to univariate time series.
Therefore, the applicability and performance of the proposed method on multivariate time series have not been explored.
Extending the shapelet discovery and matching process to multivariate time series would require channel-wise or joint multichannel shapelet discovery, together with a matching criterion that accounts for cross-channel interactions.

Another limitation is the architecture-dependent behavior observed in our experiments.
Among the three evaluated backbones, Shapelet Matching produced the largest gains with the VGG-based CNN, whose local convolutional filters are well aligned with local discriminative subsequences.
For patch-based Transformers, the gains are smaller and DBA-DTW sometimes achieves higher mean accuracy. This suggests that shapelet-based source selection may not be universally optimal across all backbone architectures.
Therefore, the proposed method should be viewed as particularly suitable for backbones with a strong local-pattern bias, rather than as a universally optimal source-selection criterion for all architectures.

Finally, the shapelet discovery hyperparameters were fixed in this study.
We used a shapelet length of 15 and 10 candidate shapelets per class for all datasets, following the previous setting, but did not conduct a sensitivity analysis over either hyperparameter.
Since the appropriate subsequence scale and number of discriminative patterns may vary across datasets, future work should explore multi-scale shapelets, validation-based hyperparameter selection, or adaptive shapelet discovery.

\section{Conclusion}
\label{chap:conc}
We introduced Shapelet Matching, a training-free source-selection method for multi-source transfer learning in time series classification. 
The method ranks candidate source datasets by comparing class-discriminative shapelets between source and target datasets, avoiding the need to pre-train a separate model for every candidate source. 
Experiments on 128 UCR datasets show that multi-source pre-training can reduce the risk of negative transfer, with the strongest gains observed for CNN-based models and competitive performance for patch-based Transformers. 
These results suggest that local shapelet-level similarity is a useful and potentially interpretable criterion for selecting source datasets when large-scale foundation-model pre-training is not available. 
This provides practitioners with a lightweight source-selection tool when large external pre-training corpora or foundation-model checkpoints are unavailable.
The main limitations are the reliance on transferable local subsequences, architecture-dependent performance, the univariate-only evaluation, and the use of fixed shapelet hyperparameters.
Future work will extend the method to multivariate time series, adaptive shapelet discovery, and architecture-aware source ranking.
\vspace{-5mm}
\section*{Declaration of Generative AI and AI-assisted Technologies in the Manuscript Preparation Process}
During the preparation of this work, the first author used Cursor for coding assistance, Grammarly for grammar checking, and Gemini and ChatGPT for wording assistance. After using these tools, the authors reviewed and edited the content as needed and take full responsibility for the content of this article.
\vspace{-5mm}
\section*{Acknowledgement}
This work was partially supported by JST BOOST, Japan Grant Number \mbox{JPMJBS2406}.
\vspace{-5mm}
\bibliographystyle{ieeetran}
\bibliography{references}

@inproceedings{Ismail_Fawaz_2018,
	doi = {10.1109/bigdata.2018.8621990},
	year = 2018,
	author = {Hassan Ismail Fawaz and Germain Forestier and Jonathan Weber and Lhassane Idoumghar and Pierre-Alain Muller},
	title = {Transfer learning for time series classification},
	booktitle = {Int. Conf. Big Data}
}

@article{iwana2021an,
  title={An Empirical Survey of Data Augmentation for Time Series Classification with Neural Networks},
  author={Iwana, Brian Kenji and Uchida, Seiichi},
  journal={PLOS ONE},
  year={2021},
  doi={10.1371/journal.pone.0254841},
}

@article{zhuang2020comprehensive,
  title={A comprehensive survey on transfer learning},
  author={Zhuang, Fuzhen and Qi, Zhiyuan and Duan, Keyu and Xi, Dongbo and Zhu, Yongchun and Zhu, Hengshu and Xiong, Hui and He, Qing},
  journal={Proc. {IEEE}},
  volume={109},
  number={1},
  pages={43--76},
  year={2021},
  publisher={IEEE}
}

@article{Lecun_1998,
	doi = {10.1109/5.726791},
	year = 1998,
	publisher = {IEEE},
	volume = {86},
	number = {11},
	pages = {2278--2324},
	author = {Y. Lecun and L. Bottou and Y. Bengio and P. Haffner},
	title = {Gradient-based learning applied to document recognition},
	journal = {Proc. {IEEE}}
}

@article{Li_2020,
	doi = {10.3390/s20154271},
	year = 2020,
	publisher = {{MDPI}},
	volume = {20},
	number = {15},
	pages = {4271},
	author = {Fr{\'{e}}d{\'{e}}ric Li and Kimiaki Shirahama and Muhammad Adeel Nisar and Xinyu Huang and Marcin Grzegorzek},
	title = {Deep Transfer Learning for Time Series Data Based on Sensor Modality Classification},
	journal = {Sensors}
}

@inproceedings{Clark_2022,
	doi = {10.1109/ccece49351.2022.9918279},
	year = 2022,
	author = {Ryan Clark and Thomas E. Doyle},
	title = {A Priori Quantification of Transfer Learning Performance on Time Series Classification for Cyber-Physical Health Systems},
	booktitle = {Canadian Conf. Electr. and Comput. Eng.}
}

@inproceedings{Gikunda_2021,
	doi = {10.1109/icmla52953.2021.00129},
	year = 2021,
	author = {Patrick Gikunda and Nicolas Jouandeau},
	title = {Homogeneous Transfer Active Learning for Time Series Classification},
	booktitle = {Int. Conf. Mach. Learn. and Appl.}
}

@article{Petitjean_2011,
	doi = {10.1016/j.patcog.2010.09.013},
	year = 2011,
	publisher = {Elsevier},
	volume = {44},
	number = {3},
	pages = {678--693},
	author = {Fran{\c{c}}ois Petitjean and Alain Ketterlin and Pierre Gan{\c{c}}arski},
	title = {A global averaging method for dynamic time warping, with applications to clustering},
	journal = {Pattern Recognit.}
}

@article{UCRArchive2018,
  title={The {UCR} time series archive},
  author={Dau, Hoang Anh and Bagnall, Anthony and Kamgar, Kaveh and Yeh, Chin-Chia Michael and Zhu, Yan and Gharghabi, Shaghayegh and Ratanamahatana, Chotirat Ann and Keogh, Eamonn},
  journal={J. Automatica Sinica},
  volume={6},
  number={6},
  pages={1293--1305},
  year={2019},
  publisher={IEEE}
}

@inproceedings{Yeh_2016,
	doi = {10.1109/icdm.2016.0179},
	year = 2016,
	author = {Chin-Chia Michael Yeh and Yan Zhu and Liudmila Ulanova and Nurjahan Begum and Yifei Ding and Hoang Anh Dau and Diego Furtado Silva and Abdullah Mueen and Eamonn Keogh},
	title = {Matrix Profile I: All Pairs Similarity Joins for Time Series: A Unifying View That Includes Motifs, Discords and Shapelets},
	booktitle = {Int. Conf. Data Mining}
}

@inproceedings{Ye_2009,
	doi = {10.1145/1557019.1557122},
	year = 2009,
	author = {Lexiang Ye and Eamonn Keogh},
	title = {Time series shapelets},
	booktitle = {Knowl. Discov. and Data Mining}
}

@InProceedings{pmlr-v119-nguyen20b,
  title = 	 {{LEEP}: A New Measure to Evaluate Transferability of Learned Representations},
  author =       {Nguyen, Cuong and Hassner, Tal and Seeger, Matthias and Archambeau, Cedric},
  booktitle = 	 {Int. Conf. Mach. Learn.},
  pages = 	 {7294--7305},
  doi = {10.5555/3524938.3525614},
  year = 	 {2020},
}

@inproceedings{Tran_2019,
	doi = {10.1109/iccv.2019.00148},
	year = 2019,
	author = {Anh Tran and Cuong Nguyen and Tal Hassner},
	title = {Transferability and Hardness of Supervised Classification Tasks},
	booktitle = {Int. Conf. Comput. Vis.}
}

@InProceedings{pmlr-v162-huang22d,
  title = 	 {Frustratingly Easy Transferability Estimation},
  author =       {Huang, Long-Kai and Huang, Junzhou and Rong, Yu and Yang, Qiang and Wei, Ying},
  booktitle = 	 {Int. Conf. Mach. Learn.},
  pages = 	 {9201--9225},
  year = 	 {2022},
}

@InProceedings{pmlr-v139-you21b,
  title = 	 {{LogME}: Practical Assessment of Pre-trained Models for Transfer Learning},
  author =       {You, Kaichao and Liu, Yong and Wang, Jianmin and Long, Mingsheng},
  booktitle = 	 {Int. Conf. Mach. Learn.},
  pages = 	 {12133--12143},
  year = 	 {2021},
}

@inproceedings{Bao_2019,
	doi = {10.1109/icip.2019.8803726},
	year = 2019,
	author = {Yajie Bao and Yang Li and Shao-Lun Huang and Lin Zhang and Lizhong Zheng and Amir Zamir and Leonidas Guibas},
	title = {An Information-Theoretic Approach to Transferability in Task Transfer Learning},
	booktitle = {Int. Conf. Image Process.}
}

@inproceedings{Yao_2010,
	doi = {10.1109/cvpr.2010.5539857},
	year = 2010,
	author = {Yi Yao and Gianfranco Doretto},
	title = {Boosting for transfer learning with multiple sources},
	booktitle = {Conf. Comput. Vis. and Pattern Recognit.}
}

@article{Huang_2012,
	doi = {10.1016/j.patrec.2011.11.023},
	year = 2012,
	publisher = {Elsevier},
	volume = {33},
	number = {5},
	pages = {568--579},
	author = {Pipei Huang and Gang Wang and Shiyin Qin},
	title = {Boosting for transfer learning from multiple data sources},
	journal = {Pattern Recognit. Lett.}
}

@inproceedings{Tan_2013,
	doi = {10.1137/1.9781611972832.27},
	year = 2013,
	author = {Ben Tan and Erheng Zhong and Evan Wei Xiang and Qiang Yang},
	title = {Multi-Transfer: Transfer Learning with Multiple Views and Multiple Sources},
	booktitle = {Int. Conf. Data Mining}
}

@article{Song_2017,
	doi = {10.1007/s00500-017-2755-8},
	year = 2018,
	publisher = {Springer },
	volume = {22},
	number = {24},
	pages = {8107--8118},
	author = {Hyun-Je Song and Seong-Bae Park},
	title = {Identifying intention posts in discussion forums using multi-instance learning and multiple sources transfer learning},
	journal = {Soft Comput.}
}

@article{Weber_2021,
	doi = {10.1109/access.2021.3134628},
	year = 2021,
	publisher = {IEEE},
	volume = {9},
	pages = {165409--165432},
	author = {Manuel Weber and Maximilian Auch and Christoph Doblander and Peter Mandl and Hans-Arno Jacobsen},
	title = {Transfer Learning With Time Series Data: A Systematic Mapping Study},
	journal = {{IEEE} Access}
}

@article{Li_2019,
	doi = {10.1109/tcyb.2019.2904052},
	year = 2020,
    volume={50},
    number={7},
	publisher = {IEEE},
	pages = {3281--3293},
	author = {Jinpeng Li and Shuang Qiu and Yuan-Yuan Shen and Cheng-Lin Liu and Huiguang He},
	title = {Multisource Transfer Learning for Cross-Subject {EEG} Emotion Recognition},
	journal = {{IEEE} Trans. Cyber.}
}

@article{SAKOE_1990,
	doi = {10.1016/b978-0-08-051584-7.50016-4},
	year = 1990,
	publisher = {Elsevier},
	pages = {159--165},
	author = {Hiroaki Sakoe and Seibi Chiba},
	title = {Dynamic Programming Algorithm Optimization for Spoken Word Recognition},
	journal = {Readings Speech Recognit.}
}

@inproceedings{Ren_2022,
	doi = {10.1109/icicml57342.2022.10009876},
	year = 2022,
	author = {Run Ren and Yameng Yang and Hailong Ren},
	title = {{EEG} Emotion Recognition using Multisource Instance Transfer Learning Framework},
	booktitle = {Int. Conf. Image Process. Comput. Vis. and Mach. Learn.}
}

@inproceedings{Dai_2007,
	doi = {10.1145/1273496.1273521},
	year = 2007,
	author = {Wenyuan Dai and Qiang Yang and Gui-Rong Xue and Yong Yu},
	title = {Boosting for transfer learning},
	booktitle = {Int. Conf. Mach. Learn.}
}

@inproceedings{simonyan2015deep,
  title={Very Deep Convolutional Networks for Large-Scale Image Recognition},
  author={Simonyan, K and Zisserman, A},
  booktitle={Int. Conf. Learn. Rep.},
  year={2015}
}

@inproceedings{
nie2022time,
title={A Time Series is Worth 64 Words:  Long-term Forecasting with Transformers},
author={Yuqi Nie and Nam H Nguyen and Phanwadee Sinthong and Jayant Kalagnanam},
booktitle={Int. Conf. Learn. Rep.},
year={2023}
}

@inproceedings{lee2024model,
  title={Model Selection with a Shapelet-Based Distance Measure for Multi-source Transfer Learning in Time Series Classification},
  author={Lee, Jiseok and Iwana, Brian Kenji},
  booktitle={Int. Conf. Pattern Recognit.},
  pages={160--175},
  year={2025},
  organization={Springer},
  doi={10.1007/978-3-031-78398-2_11},
}

@inproceedings{
dosovitskiy2020image,
title={An Image is Worth 16x16 Words: Transformers for Image Recognition at Scale},
author={Alexey Dosovitskiy and Lucas Beyer and Alexander Kolesnikov and Dirk Weissenborn and Xiaohua Zhai and Thomas Unterthiner and Mostafa Dehghani and Matthias Minderer and Georg Heigold and Sylvain Gelly and Jakob Uszkoreit and Neil Houlsby},
booktitle={Int. Conf. Learn. Rep.},
year={2021},
}

@inproceedings{xu2023fast,
  title={Fast and accurate transferability measurement by evaluating intra-class feature variance},
  author={Xu, Huiwen and Kang, U},
  booktitle={Int. Conf. Comput. Vis.},
  pages={11474--11482},
  year={2023}
}

@article{Bozinovski_2020, 
 title={Reminder of the First Paper on Transfer Learning in Neural Networks, 1976}, 
 volume={44}, 
DOI={10.31449/inf.v44i3.2828}, 
number={3}, 
journal={Informatica}, 
publisher={Slovenian Association Informatika}, 
author={Bozinovski, Stevo}, 
year={2020}}

@article{zhang2022survey,
  title={A survey on negative transfer},
  author={Zhang, Wen and Deng, Lingfei and Zhang, Lei and Wu, Dongrui},
  journal={J. Automatica Sinica},
  volume={10},
  number={2},
  pages={305--329},
  year={2023},
  publisher={IEEE}
}

@inproceedings{liu2024timer,
  title={Timer: generative pre-trained transformers are large time series models},
  author={Liu, Yong and Zhang, Haoran and Li, Chenyu and Huang, Xiangdong and Wang, Jianmin and Long, Mingsheng},
  booktitle={Int. Conf. Mach. Learn.},
  pages={32369--32399},
  year={2024}
}

@article{zhou2023one,
  title={One fits all: Power general time series analysis by pretrained lm},
  author={Zhou, Tian and Niu, Peisong and Sun, Liang and Jin, Rong and others},
  journal={Adv. Neural Inf. Process. Sys.},
  volume={36},
  pages={43322--43355},
  year={2023}
}

@inproceedings{goswami2024moment,
  title={{MOMENT}: A Family of Open Time-series Foundation Models},
  author={Goswami, Mononito and Szafer, Konrad and Choudhry, Arjun and Cai, Yifu and Li, Shuo and Dubrawski, Artur},
  booktitle={Int. Conf. Mach. Learn.},
  pages={16115--16152},
  year={2024},
}

@article{gao2024units,
  title={{UniTS}: A unified multi-task time series model},
  author={Gao, Shanghua and Koker, Teddy and Queen, Owen and Hartvigsen, Thomas and Tsiligkaridis, Theodoros and Zitnik, Marinka},
  journal={Adv. Neural Inf. Process. Sys.},
  volume={37},
  pages={140589--140631},
  year={2024}
}

@article{ragab2022self,
  title={Self-supervised autoregressive domain adaptation for time series data},
  author={Ragab, Mohamed and Eldele, Emadeldeen and Chen, Zhenghua and Wu, Min and Kwoh, Chee-Keong and Li, Xiaoli},
  journal={IEEE Trans. Neural Netw. and Learn. Syst.},
  volume={35},
  number={1},
  pages={1341--1351},
  year={2024},
  publisher={IEEE}
}

@article{deng2024domain,
  title={Domain generalization in time series forecasting},
  author={Deng, Songgaojun and Sprangers, Olivier and Li, Ming and Schelter, Sebastian and De Rijke, Maarten},
  journal={ACM Trans. Knowl. Discov. from Data},
  volume={18},
  number={5},
  pages={1--24},
  year={2024},
  publisher={ACM}
}

@InProceedings{liu2025learning,
  title = 	 {Learning Soft Sparse Shapes for Efficient Time-Series Classification},
  author =       {Liu, Zhen and Luo, Yicheng and Li, Boyuan and Eldele, Emadeldeen and Wu, Min and Ma, Qianli},
  booktitle = 	 {Int. Conf. Mach. Learn.},
  pages = 	 {39032--39059},
  year = 	 {2025},
  volume = 	 {267},
  publisher =    {PMLR}
}

@article{liu2026unishape, 
title={A Unified Shape-Aware Foundation Model for Time Series Classification}, 
volume={40},
number={28},
journal={AAAI Conf. Artif. Intell.},
author={Liu, Zhen and Wang, Yucheng and Li, Boyuan and Zheng, Junhao and Eldele, Emadeldeen and Wu, Min and Ma, Qianli},
year={2026}, 
pages={23972--23980}}
\end{document}